\documentclass{article}

\usepackage{arxiv_sty}

\usepackage[utf8]{inputenc} 
\usepackage[T1]{fontenc}    
\usepackage{hyperref}       
\usepackage{url}            
\usepackage{booktabs}       
\usepackage{amsfonts}       
\usepackage{nicefrac}       
\usepackage{microtype}      
\usepackage{xcolor}         
\usepackage{amsmath}
\usepackage{amssymb}
\usepackage{mathtools}
\usepackage{amsthm}
\usepackage{multirow}
\usepackage{float}
\usepackage{algorithm}
\usepackage{algorithmic}
\usepackage{graphicx}
\usepackage{subfig}
\usepackage[export]{adjustbox}
\usepackage{caption}
\usepackage{wrapfig}
\usepackage{setspace}
\usepackage[textsize=tiny]{todonotes}

\usepackage{caption}
\title{Heterogenous Memory Augmented Neural Networks}

\author{%
  Zihan Qiu\thanks{Work done while interning at HKUST and BAAI: qzh11628@gmail.com} \\
  Tsinghua University\\
  \And
  Zhen Liu \\
  University of Montr\'eal \\
  \AND
  Shuicheng Yan \\
  BAAI \\
  \And
  Shanghang Zhang \\
  Peking University \\
  \And
  Jie Fu\thanks{Corresponding author: jiefu@ust.hk} \\
  HKUST 
}

\begin{document}
\maketitle
\begin{abstract}
It has been shown that semi-parametric methods, which combine standard neural networks with non-parametric components such as external memory modules and data retrieval, are particularly helpful in data scarcity and out-of-distribution (OOD) scenarios. 
However, existing semi-parametric methods mostly depend on independent raw data points - this strategy is difficult to scale up due to both high computational costs and the incapacity of current attention mechanisms with a large number of tokens.
In this paper, we introduce a novel heterogeneous memory augmentation approach for neural networks which, by introducing learnable memory tokens with attention mechanism, can effectively boost performance without huge computational overhead. Our general-purpose method can be seamlessly combined with various backbones (MLP, CNN, GNN, and Transformer) in a plug-and-play manner.
We extensively evaluate our approach on various image and graph-based tasks under both in-distribution (ID) and OOD conditions and show its competitive performance against task-specific state-of-the-art methods. Code is available at \url{https://github.com/qiuzh20/HMA}.

\end{abstract}

\section{Introduction}

Semi-parametric methods, which parametrize the mapping from an input domain $\mathcal{X}$ to an output domain $\mathcal{Y}$ with both neural net parameters and non-parametric data, are widely used in a variety of tasks including but not limited to meta-learning~\cite{munkhdalai2017meta, snell2017prototypical}, energy-based models~\cite{arbel2020generalized} and planning~\cite{humphreys2022large}. With a non-parametric component in the neural net architecture, the model may better incorporate priors like distances between data points and characterize attributes such as data uncertainty. A typical design is to retrieve data associated with the current input with k-nearest neighbors (kNN). For instance, kNN-LM~\cite{khandelwal2019generalization} proposes to perform inference with data retrieval from a small batch of inputs, similar to the concept of support set used in few-shot learning~\cite{sprechmann2018memory}. In practice, the retrieval strategies and parameters, such as the number of nearest neighbors, must be carefully designed for performance and low computational overhead~\cite{drozdov2022you, peng2023semiparametric}.

Instead of using full datasets, one may augment the neural net architecture with \emph{external} memory. With an external memory module, one can store a tiny amount of data points or latent features obtained through dataset distillation~\cite{wang2018dataset}, random selection~\cite{kossen2021self}, or learned policies~\cite{peng2023semiparametric}. Especially with the attention mechanism employed by NPT~\cite{kossen2021self}, a model with external memory can utilize non-parametric components more flexibly than traditional methods such as Gaussian processes~\cite{damianou2013deep} due to the learnable dependencies instead of a fixed kernel function. Still, it suffers from the same scalability problem as the attention has to operate directly on a potentially large data set or features.

\begin{figure}
  \begin{center}
  \centerline{\includegraphics[width=0.9\columnwidth]{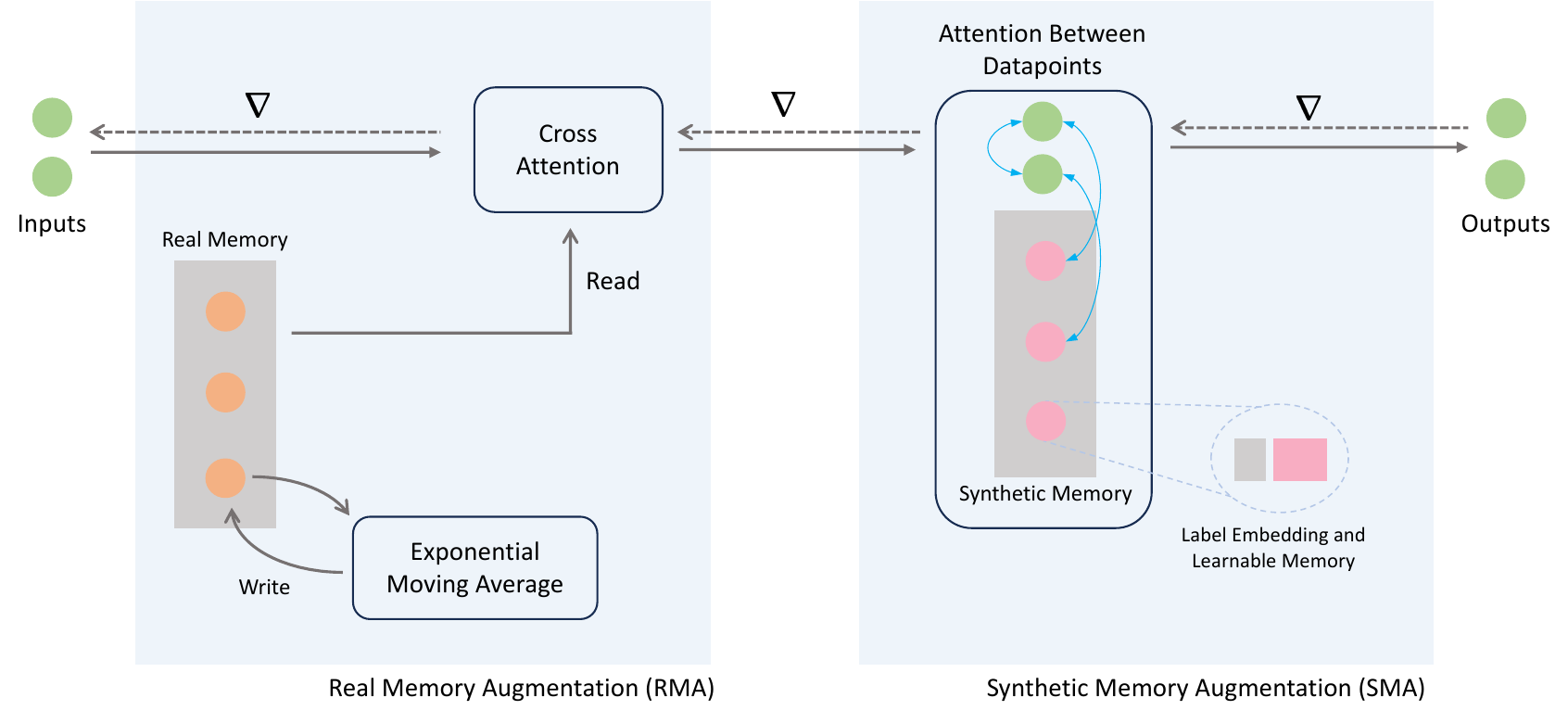}}
  \vskip -0.1in
  \caption{\ The overall framework of HMA. For inputs from some backbone, we concatenate them with label embeddings from \textbf{known} token similar to CLS. In RMA, a momentum memory queue and attention-based reader provide inputs with cross-batch information. During training, the real memory is updated with features from the momentum encoder and \textbf{true} label embeddings. In SMA, attention between datapoints is used to pass information among inputs and synthetic memory slots.}
  \vskip -0.15in
  \label{HMA}
  \end{center}
\end{figure}

Inspired by some recent work that effectively learns a tiny amount of synthetic data for teaching student networks ~\cite{wang2018dataset,zhao2023dataset}, we propose Heterogeneous Memory Augmentation (HMA), a general approach for learning dependencies between data and augmenting networks without high computational overhead for various downstream tasks.
HMA sequentially employs on the feature space of some backbone network (a) a real memory augmentation module (RMA), with classical memory augmentation methods following~\cite{munkhdalai2017meta, he2020momentum}, and (b) a synthetic memory augmentation module (SMA).
Specifically, SMA learns compressed memory entries that encode \emph{dataset-relevant} information and leverages attention across datapoints (ABD) which consider the cross-relationship between the input batch in a manner similar to NPT.
These learned memory entries assist SMA in working effectively without relying on large batch sizes or additional selection methods.
Notably, our HMA is architecture-agnostic and can be plugged into almost any backbone architecture in a few lines of code, without any additional training phases or losses.

In summary, our contributions are:
\begin{itemize} 
    \item We introduce heterogeneous memory augmentation (HMA), a better semi-parametric module that leverages both synthetic (learnable) and real (past data) memory. Compared to previous methods, our HMA can better capture the data dependencies for semi-parametric inference.
    \item The key component, synthetic memory augmentation (SMA), utilizes attention mechanisms to perform non-parametric inference on top of learned memory slots and inputs. The use of attention across datapoints allows the module to better capture the data dependencies 1) between the input batch and the memory slots as well as 2) between individual datapoints in the input batch, compared to traditional non-parametric techniques like Gaussian processes.
    \item Our model-agnostic HMA directly operates on the feature space and, therefore, can be incorporated into various backbones in a plug-in-play fashion. 
    \item With learnable memory slots in SMA, HMA is more tailored to downstream tasks. Extensive experiments show its effectiveness in various in-distribution and out-of-distribution tasks.
\end{itemize}

\section{Related Work}
\vskip -0.1in
\subsection{Semi-parametric Methods}
\vskip -0.1in
Retrieval methods leverage the similarity between data points to provide information for prediction.
For input $x$, the prediction probability is $p=\lambda p_{\text{kNN}}+(1-\lambda)p_{\text{BaseModel}}$ to improve its ability to handle long-tail data and robustness~\cite{khandelwal2019generalization}.
During testing, the model can directly utilize similar training data for predictions.
These characteristics make the retrieval method bring substantial improvements in various tasks of NLP~\cite{karpukhin2020dense,xu2022boosting,lin2021bertgcn} and graph tasks~\cite{Wang2022RetrievalenhancedGN}.
GNN-LM~\cite{meng2021gnn} also builds a graph by combining the samples with their corresponding retrieval data and further enhances the sample features through graph aggregation.
Non-Parametric Classifier~\cite{orhan2018simple, wu2018unsupervised} stores the training data's features and labels, enhancing similarity-based classification.
Prototypical Networks~\cite{snell2017prototypical, pahde2021multimodal, sun2019hierarchical} employ specifically designed objectives to learn a metric space. Each class's non-parametric "prototype representations" are easily obtained and used for few-shot and zero-shot tasks in this space. Compared to these methods, HMA learns synthetic memory slots directly with gradient descent updates, which can be more tailored for downstream tasks.

\subsection{Attention}
\vskip -0.1in
Attention~\cite{vaswani2017attention} can integrate various information and offer a versatile mechanism to uncover and exploit relationships between data.
BatchFormer~\cite{hou2022batchformer} performs attention in batches, allowing the model weights to learn the relationships.
NPT~\cite{kossen2021self} deploys Attention between Attributes and Attention between Datapoints to incorporate the learning of representation and relationships between data.
However, ABD in NPT leads to a computational complexity of $O(n^2)$, which makes it difficult to incorporate a large amount of data and further hinders the feature learning process.
To address the efficiency issue, \cite{rastogi2022semi} presents inducing points to avoid expensive full self-attention.

\subsection{Neural Memory}
\vskip -0.1in
Memory mechanisms can be used to discover and exploit the relationships between data through contrastive learning \cite{he2020momentum,qu2020coda}.
Such relationships are also pervasive in meta-learning.
\cite{wu2021learning} proposes a learnable memory to capture the meta-class information of semantic segmentation during the base class training procedure.
Unlike the works mentioned above that use memory in a no-parametric fashion, \cite{sprechmann2018memory} utilizes relevant features in the input to recall memory and uses them to modify the network parameters. 
Meta Networks \cite{munkhdalai2017meta} and Neural Stored-program Memory~\cite{le2019neural} directly store network parameters in different training stages in memory.

Soft prompts~\cite{liu2021gpt, liu2021p, sandler2022fine} can also be regarded as a form of learnable memory that captures task-relevant information during fine-tuning and aids in leveraging the knowledge within pre-trained models. L2P~\cite{wang2022learning, wang2022dualprompt} stores a series of soft prompts and reads them using key-value matching, resulting in remarkable performance during continued learning.

\vskip -0.2in
\section{Preliminaries}
\vskip -0.1in
\textbf{Attention mechanism} \cite{vaswani2017attention} is a fundamental component in deep learning models that enables the model to focus on relevant information selectively. It is commonly formulated as follows:
\begin{small}
\begin{equation}
\begin{aligned}
\text{Att}(Q, K, V)=\text{softmax}(QK^T/\sqrt{d_k})V,
\end{aligned}
\end{equation}
\end{small}
where $Q, K, V$ are stacked matrixes of queries, keys, and values. Typically, multi-head attention (MultiHA) is used for large model capacity. Using multiple attention modules in parallel, the model can capture different aspects of the input data and combine their results. The outputs of the attention heads are concatenated to yield the final output:
\begin{equation}
\text{MultiHA}(Q, K, V)=\text{concat}(\text{head}_1, \cdots, \text{head}_h), \text{ where }\text{head}_i=\text{Att}(QW_i^Q, KW_i^K, VW_i^V)
\nonumber 
\end{equation}

where $W_i^{Q},W_{i}^K$ and $W_{i}^V$ are learnable matrices for the $d_k$-dimensional queries, keys and values for $i$-th head. By $AB$ with $A \in \mathbb{R}^{n \times a \times b}$ and $B \in \mathbb{R}^{n \times b \times c}$, we mean batched matrix multiplication in Einstein summation notation $\text{einsum}(A, B, \text{'nab,nbc->nac'})$. The output of MHA is computed with additional components, including feed-forward layer (FF), layer normalization (LN) and residual connection:
\begin{small}
\begin{equation}
\label{MHA}
\text{MHA}(Q, K, V)=H + \text{FF}(\text{LN}(H)), \text{ where } H=V+\text{MultiHA}(\text{LN}(Q), \text{LN}(K), \text{LN}(V))
\end{equation}
\end{small}
\vskip -0.1in
\textbf{Non-parametric transformer (NPT)} \cite{kossen2021self} extends the standard transformer in a semi-parametric way by leveraging the dependencies between datapoints with the attention among large batched inputs. 
Specifically, NPT first encodes the input batch $H \in \mathbb{R}^{bs \times d}$ and reshape the output into $H' = \mathbb{R}^{1 \times bs \times (d*h)}$, and then applies attention across datapoints (ABD) by treating the input batch dimension as the token dimension:
\vskip -0.1in
\begin{equation}
\label{ABD}
H_{\text{ABD}}' = \text{ABD}(H) = \text{MHA}(H', H', H') \in \mathbb{R}^{1 \times bs \times (d*h)}
\end{equation}
\vskip -0.1in
Once information has been exchanged between datapoints in the input batch, the ABD output $H_{\text{ABD}}' \in \mathbb{R}^{bs \times 1 \times (d*h)}$ is reshaped back into $H_{\text{ABD}} \in \mathbb{R}^{bs \times d \times h}$ to apply a self-attention operation on the feature dimension $d$, which we term attention across attributes (ABA) to distinguish it from ABD:

\begin{equation}
\begin{aligned}
\label{ABA}
\text{ABA}(H_{\text{ABD}})=\text{MHA}(H_{\text{ABD}}, H_{\text{ABD}}, H_{\text{ABD}})\in \mathbb{R}^{bs \times d \times h}
\end{aligned}
\end{equation}

ABD and ABA can be iteratively applied for several times before feeding the resulted (and reshaped) feature into a classifier or some other neural net.

\section{Heterogenous Memory Augmentation}
\vskip -0.1in
HMA performs two consecutive steps on the extracted feature batch~\ref{3.1}: 1) real memory augmentation (RMA)~\ref{3.2}, and 2) synthetic memory augmentation (SMA)~\ref{3.3}.
As semi-parametric methods heavily rely on the quality of non-parametric components, we employ RMA to \emph{expand the non-parametric part implicitly}. Subsequently, we combine it with synthetic memory and apply attention between datapoints.
The final output feature from SMA is fed into a linear projection head to compute logits for classification. We summarize HMA in the supplementary material.

\subsection{Feature Extraction}
\label{3.1}
\vskip -0.1in
Given input $X$ and label $Y$ with batch size $bs$, We compute the feature $h^1 = E(X) \in \mathbb{R}^{bs\times d_1}$, where $E(\cdot)$ can be any backbone, and $d_1, d_2$ is the feature embedding dimension.
In line with the setup of NPT, we incorporate labels as attributes. However, directly adding the embeddings of the true labels would leak information.
To address this, for an n-classification problem, we replace the true labels $Y$ with an additional label, denoted as n+1, and treat it as the CLS token as common in transformer literature \cite{devlin2018bert}. 
We compute label embedding $e = E_{\text{label}}(\textbf{n+1})\in \mathbb{R}^{bs \times d_2}$, where $d_2$ is the dimension of the label embedding. After concatenating, we obtain aggregated features $C^1 = [h^1; e] \in \mathbb{R}^{bs \times d}, d = d_1 + d_2$. This concatenated representation $C^1$ is used for later processes.

\subsection{Real Memory Augmentation (RMA)}
\label{3.2}
\vskip -0.1in
The properties of non-parametric components greatly influence the effectiveness of semi-parametric methods.
NPT uses large batch sizes to reduce the influence of randomly selected non-parametric components.
Retrievers ensure reference data quality through top-k selection. We aim to develop a method that effectively provides information beyond individual batches.

Drawing inspiration from MoCo~\cite{he2020momentum}, which employs a momentum-updated memory queue to provide stable and extensive contrastive samples, we introduce a real memory buffer $M_{\text{buffer}}\in \mathbb{R}^{m_1\times d}$ with size $m_1$ (this buffer is empty at the beginning) and a momentum encoder $E_{m}$.

$\textbf{Reading.}$ We utilize MHA\ref{MHA} to extract information from the memory buffer, thereby implicitly providing cross-batch information for subsequent SMA. To achieve this, we first transform $M_{\text{buffer}}$ from $\mathbb{R}^{m_1\times d}$ to $\mathbb{R}^{1 \times m_1\times d}$ and $C^1$ from $\mathbb{R}^{bs \times d}$ to $\mathbb{R}^{bs\times 1 \times d}$. Then, we repeat $M_{\text{buffer}}$ $bs$ times in the first dimension to obtain $M_{\text{buffer}}'\in \mathbb{R}^{bs \times m_1\times d}$. Subsequently, we concatenate $M_{\text{buffer}}'$ and $C^1$ along the second dimension, resulting in $C^1_{\text{aug}}\in \mathbb{R}^{bs\times 1+m_1\times d)}$. RMA computes the output feature $C^2$ by:
\begin{equation}
\label{eqn:RMA}
C^2=\text{MHA}(C^1, C_{\text{aug}}^{1},C^1_{\text{aug}})\in \mathbb{R}^{bs\times 1 \times d}
\end{equation}
\vskip -0.05in
After RMA, $C^2$ incorporates information from real data beyond a single batch.
In \ref{memory properties}, we observed that using RMA alone seldom brings improvements. Moreover, while using SMA alone can lead to some improvements, adding RMA achieves pronounced improvements.

$\textbf{Writing.}$  We insert new aggregated input features $[E_{m}(X), E_{\text{label}}(Y)]\in \mathbb{R}^{bs\times d}$ into the memory queue, and update the momentum encoder momentum update rule as in MoCo: $E_{m}^{t+1}=\lambda E_{m}^t + (1 - \lambda) E^{t+1}$, in which $E^{t+1}$ is the SGD-updated encoder from $E^{t}$. 

\subsection{Synthetic Memory Augmentation (SMA)}
\vskip -0.1in
In NPT, the ABD mechanism allows models to utilize non-parametric components flexibly. 
We can also leverage this mechanism to introduce learnable synthetic memory, which captures dataset-relevant information. These synthetic memories also serve as compact non-parametric components for the ABD mechanism, eliminating the need for NPT's reliance on large batch sizes.

Although we can directly initialize a series of learnable parameters as memory slots with dimensions $d_1+d_2$, we also aim to incorporate label information into the inference process of SMA. 
Specifically, for each category $i \in [n]$, we initialize $m_2$ learnable memory slots.
Each memory slot is in $\mathbb{R}^{d_1}$ and has corresponding label $y_i$.
With the label embedder $E_{\text{emb}}$, we map the label $y_i$ to the label embedding $e_{\text{syn}, i} \in \mathbb{R}^{d_2}$.
By concatenating all learnable memory slots and label embeddings, we obtain a learnable memory $M_{\text{SMA}} \in \mathbb{R}^{(n * m_2) \times (d_1 + d_2)}$. 
We reshape $C^2$ and $M_{\text{SMA}}$, and concatenate them to obtain $C^2_{\text{aug}} \in \mathbb{R}^{1\times (bs+m_2*n)\times d}$. The SMA output features are obtained by
\begin{equation}
\label{eqn:SMA}
C^3=\text{MHA}(C^2, C^2_{\text{aug}}, C^2_{\text{aug}})\in \mathbb{R}^{1\times bs\times d}.
\end{equation}

Finally, we use a linear projection head to compute the logits from $C^3$. During the training phase, the learnable memory slots are updated by gradient, while the label embeddings are regenerated using the updated embedder $E_{\text{emb}}$.

\section{Experiments}
\vskip -0.1in
In this section, we evaluate the performance of HMA and explore its characteristics. 
Section~\ref{general} demonstrates that HMA can be combined with various backbones and improve image and graph tasks.
Following the success of semi-parametric methods on OOD tasks, in section~\ref{ood-perform}, we examine HMA's out-of-distribution generalization capabilities under different distributional shifts.
Finally, in Section~\ref{memory properties}, we demonstrate the characteristics of both real and synthetic memory slots.
Ablation study shows while ABD itself can bring improvements, with the help of real memory and synthetic memory slots, HMA consistently achieves competitive performance against task-specific state-of-the-art methods. 

\textbf{Experiment Setup.} Our experiments strictly follow the basic settings like batch size, learning rate schedule, number of epochs, optimizer, evaluation metric, and test model selection in previous works for a fair comparison.
In CIFAR-10 task, we use the codebase\footnote{https://github.com/kuangliu/pytorch-cifar}.
Each method is trained 200 epochs, and the final test accuracies over three random seeds are reported.
In tuning ViT \cite{dosovitskiy2020image} task, we follow \cite{sandler2022fine} to tune the ViT-B/32 model, which was pretrained on Imagenet-21K \cite{deng2009imagenet}.
In Colored MNIST task \ref{CMNIST}, we use code base from \cite{arjovsky2019invariant}.
In obg molecular property prediction tasks \ref{ogb}, we use the official code base.
In real-world graph ID tasks \ref{GraphID}  and graph OOD tasks \ref{OOD-mat}\ref{OOD-SST}, we follow the official implementation of G-mixup \cite{wang2021mixup} and CIGA \cite{han2022g}.
More details can be found in the \underline{supplementary material}. 

\textbf{Table Notations.}\label{label notations} We have \textbf{bolded} the best results and added an \underline{underline} to the second-best results in all the tables. 
To present the effects of different components in HMA and provide a comprehensive view, we included the results of ablations for each component in the subsequent tables: 1) RMA represents only real memory augmentation from \ref{3.2}. 2) ABD and 3) ABD+SYN represent the use of only attention between datapoints from \ref{3.3} without and with learnable synthetic memory, respectively. 4) ABD+RMA represents using attention between datapoints and RMA.

\subsection{HMA is a General Augmentation Mechanism}
\label{general}
\vskip -0.1in
We initially tested HMA on CIFAR-10. We adopted a ResNet-18~\cite{he2016deep} encoder, similar to the one used in NPT~\cite{kossen2021self}.
The original NPT framework reported a performance of 93.7\% on CIFAR-10, which is even lower than using ResNet alone (93.9\%).
HMA achieves an accuracy of 95.54±0.11\%, slightly outperforming the backbone ResNet (which achieves 95.43±0.01\%). 
We acknowledge that minor differences may be attributed to the fact that this setting has already been well-studied. 

To evaluate the performance of HMA on larger models and its adaptability to pre-trained models, we conducted experiments following the setup of ViT~\cite{dosovitskiy2020image} on the pets37~\cite{parkhi12a}, flowers102~\cite{Nilsback08}, CIFAR100~\cite{krizhevsky2009learning}, and DTD~\cite{cimpoi14describing} datasets.
We performed a simple hyperparameter search for HMA using different values of $m_1={1,2,4}\times bs$ and $m_2 = {4,8,16}$.
In addition to directly tuning the ViT backbone, we compared HMA with the SOTA methods \cite{sandler2022fine} that introduced soft prompts (referred to as learnable memory in the paper).
"prompt-p" in the table indicates adding soft prompts only at the input layer, similar to p-tuning \cite{liu2021gpt}. "prompt-a" refers to independently adding soft prompts at each layer of ViT, similar to p-tuning v2~\cite{liu2021p}. 
PARAM refers to the ablation where non-parametric memory is not introduced, and only the parameters of the MHA~\ref{MHA} module are used. We report the mean and standard deviation for 3 seeds. The results are summarized in \tablename~\ref{ViT-full}.

\begin{table*}[thb]
\caption{\ Accuracy of full fine-tuning pretrained ViT on 4 image datasets. PROMPT-P and PROMPT-A indicate adding soft prompts only at the input layer and at each layer, respectively. PARAM is the parameter ablation. Other notations are introduced in~\ref{label notations}.}
\begin{center}
\begin{small}
\begin{sc}
\vskip -0.1in
\resizebox{0.8\columnwidth}{!}{
\begin{tabular}{lcccccc}
\toprule
dataset  &\scalebox{0.8}{DTD} &\scalebox{0.8}{PETS37}  &\scalebox{0.8}{FLOWERS102}  &\scalebox{0.8}{CIFAR100}  &\scalebox{0.8}{AVG}     \\
\midrule
vit       &75.11\scalebox{0.8}{$\pm$0.24}       &\underline{90.71}\scalebox{0.8}{$\pm$0.07}     &98.48\scalebox{0.8}{$\pm$0.40}       &91.48\scalebox{0.8}{$\pm$0.27}    &88.95       \\
\scalebox{0.8}{prompt-p}     &75.69\scalebox{0.8}{$\pm$0.56}    &90.49\scalebox{0.8}{$\pm$0.22}     &98.59\scalebox{0.8}{$\pm$0.22}       &91.61\scalebox{0.8}{$\pm$0.14}    &89.09     \\
\scalebox{0.8}{prompt-a}     &75.35\scalebox{0.8}{$\pm$0.81}    &90.47\scalebox{0.8}{$\pm$0.28}      &98.42\scalebox{0.8}{$\pm$0.11}      &91.62\scalebox{0.8}{$\pm$0.07}    &88.97     \\
\midrule
\scalebox{0.8}{param}     &75.12\scalebox{0.8}{$\pm$0.67}       &90.27\scalebox{0.8}{$\pm$0.05}      &98.46\scalebox{0.8}{$\pm$0.29}      &\underline{91.69}\scalebox{0.8}{$\pm$0.08}    &88.89     \\
rma       &75.67\scalebox{0.8}{$\pm$0.16}        &89/96\scalebox{0.8}{$\pm$0.31}      &98.61\scalebox{0.8}{$\pm$0.05}      &91.63\scalebox{0.8}{$\pm$0.31}    &88.97   \\
abd       &\underline{75.89}\scalebox{0.8}{$\pm$1.04}       &90.66\scalebox{0.8}{$\pm$0.20}      &98.57\scalebox{0.8}{$\pm$0.05}       &91.65\scalebox{0.8}{$\pm$0.14}    &\underline{89.19}   \\
\scalebox{0.8}{abd+rma}   &75.69\scalebox{0.8}{$\pm$0.40}       &90.54\scalebox{0.8}{$\pm$0.77}      &\underline{98.69}\scalebox{0.8}{$\pm$0.14}       &91.63\scalebox{0.8}{$\pm$0.19}    &89.14   \\
\scalebox{0.8}{abd+syn}   &75.69\scalebox{0.8}{$\pm$0.75}       &90.09\scalebox{0.8}{$\pm$1.10}      &98.49\scalebox{0.8}{$\pm$0.15}       &91.72\scalebox{0.8}{$\pm$0.08}    &89.00   \\
hma  &\textbf{76.05}\scalebox{0.8}{$\pm$0.39}  &\textbf{91.08}\scalebox{0.8}{$\pm$0.22}     &\textbf{98.70}\scalebox{0.8}{$\pm$0.16}   &\textbf{91.93}\scalebox{0.8}{$\pm$0.19}  &\textbf{89.44}\\
\bottomrule
\end{tabular}
}
\end{sc}
\end{small}
\end{center}
\label{ViT-full}
\end{table*}

From \tablename~\ref{ViT-full}, we can observe that the complete HMA consistently achieves better results than the best prompt tuning. The inferior performance of prompt tuning may be attributed to the interference caused by inserting new randomly initialized parameters. In contrast, HMA operates solely in the feature space of the backbone's output, avoiding such interference and allowing for more effective utilization of the original parameters.

We also conducted experiments following \cite{sandler2022fine} in the scenario where only the CLS token and head of the pretrained model are tuned, and the results are shown in \tablename~\ref{ViT-part-combine}. Since HMA and prompt tuning are orthogonal techniques, we also examined the effects of adding HMA on top of prompt tuning. Aloughth HMA, operating solely in the feature space, does not show as significant improvement as prompt tuning when most of the model parameters are fixed, it can be combined with prompt tuning to achieve more pronounced improvements.
\vskip -0.1in
\begin{table*}[hbt]
\caption{\ Accuracy of only tuning CLS token and head of ViT. +HMA indicates adding HMA to the original method. The results on the right side of $\rightarrow$ represent the performance after adding HMA.}
\begin{center}
\begin{sc}
\vskip -0.1in
\begin{small}
\resizebox{0.8\columnwidth}{!}{
\begin{tabular}{ccccccc}
\toprule
dataset     & \scalebox{0.8}{DTD}          & \scalebox{0.8}{PETS37}              \\
\midrule
vit$\rightarrow$+hma      &74.01\scalebox{0.8}{$\pm$0.29}$\rightarrow$74.91\scalebox{0.8}{$\pm$0.39}    &89.17\scalebox{0.8}{$\pm$0.07}$\rightarrow$89.72\scalebox{0.8}{$\pm$0.22} \\
prompt-p$\rightarrow$+hma   &73.98\scalebox{0.8}{$\pm$0.56}$\rightarrow$ 73.95\scalebox{0.8}{$\pm$0.73}   &89.22\scalebox{0.8}{$\pm$0.22}$\rightarrow$ 89.58\scalebox{0.8}{$\pm$0.60}  \\
prompt-a$\rightarrow$+hma   &\underline{74.75}\scalebox{0.8}{$\pm$0.86}$\rightarrow$\textbf{75.44}\scalebox{0.8}{$\pm$0.06}  &\underline{89.88}\scalebox{0.8}{$\pm$0.28}$\rightarrow$\textbf{90.71}\scalebox{0.8}{$\pm$0.27} \\
\bottomrule
\end{tabular}
}
\end{small}
\end{sc}
\end{center}
\vskip -0.1in
\label{ViT-part-combine}
\end{table*}
To further validate the effectiveness of HMA with diverse backbones, we combine HMA with GCN \cite{kipf2016semi}, GIN \cite{xu2018powerful} and their variants GCN-v, GIN-v\cite{gilmer2017neural, li2017learning} backbones on three molecular property prediction tasks on the ogb benchmark \cite{hu2020open}. The results are summarized in \tablename~\ref{ogb}. 
HMA brings improvement among 10 of 12 reported AUROCs.
\begin{table}[hbt]
\caption{\ AUROC and standard deviation of backbones and HMA on ogb benchmark. We have marked the relative improvements over the backbone with an upward arrow ($\uparrow$) for the average results.}
\begin{center}
\begin{sc}
\resizebox{0.7\columnwidth}{!}{
\begin{tabular}{lcccc}
\toprule
Dataset & bace & bbbp & hiv  & avg\\
\midrule
GCN    & 79.15\scalebox{0.8}{$\pm$1.44}&   68.87\scalebox{0.8}{$\pm$1.51}&  76.06\scalebox{0.8}{$\pm$0.97} & 74.69 \\
\scalebox{0.8}{GCN+HMA} & 80.59\scalebox{0.8}{$\pm$1.15}&   69.96\scalebox{0.8}{$\pm$ 0.5}5& 77.26\scalebox{0.8}{$\pm$1.42} & 75.85($\uparrow$1.16)\\
\midrule
GCN-v    & 68.93\scalebox{0.8}{$\pm$6.95}&   67.80\scalebox{0.8}{$\pm$2.35}& 75.99\scalebox{0.8}{$\pm$1.19} & 70.91\\
\scalebox{0.8}{GCN-v+HMA} & 72.53\scalebox{0.8}{$\pm$1.36}&   68.26\scalebox{0.8}{$\pm$0.85}&   76.51\scalebox{0.8}{$\pm$1.70} & 72.73($\uparrow$1.82)      \\
\midrule
GIN     & 72.97\scalebox{0.8}{$\pm$4.00}&   68.17\scalebox{0.8}{$\pm$1.48}& 75.58\scalebox{0.8}{$\pm$1.40} & 72.24\\
\scalebox{0.8}{GIN+HMA} & 77.18\scalebox{0.8}{$\pm$2.20}& 69.47\scalebox{0.8}{$\pm$1.50}& 76.47\scalebox{0.8}{$\pm$0.17}  & 74.38($\uparrow$2.14)\\
\midrule
GIN-v      & 73.46 \scalebox{0.8}{$\pm$5.24} & 69.71\scalebox{0.8}{$\pm$1.92}&   77.07\scalebox{0.8}{$\pm$1.40}   &73.41   \\
\scalebox{0.8}{GIN-v+HMA} & 76.56\scalebox{0.8}{$\pm$1.38} & 68.43\scalebox{0.8}{$\pm$1.19}& 75.83\scalebox{0.8}{$\pm$1.23} &73.61($\uparrow$0.20)\\
\bottomrule
\end{tabular}
}
\end{sc}
\end{center}
\vskip -0.1in
\label{ogb}
\end{table}

\begin{table*}[thb]
\caption{\ Accuarcies of different augmentation methods on real-world graphs. We strictly follow the official implementation and report the baseline results from G-mixup \cite{han2022g}.}
\begin{center}
\begin{sc}
\vskip -0.1in
\begin{small}
\resizebox{0.9\columnwidth}{!}{
\begin{tabular}{lcccccc}
\toprule
Dataset     & IMDB-B                     & IMDB-M            & REDD-B            & REDD-M5     & REDD-M12     &avg     \\
\midrule
vanilla       & 71.55\scalebox{0.8}{$\pm$3.53}                 & 48.83\scalebox{0.8}{$\pm$2.75}        & 92.59\scalebox{0.8}{$\pm$0.86}        & 55.19\scalebox{0.8}{$\pm$1.02}   & 50.23\scalebox{0.8}{$\pm$0.83}  &63.68       \\
D-edge     & 72.20\scalebox{0.8}{$\pm$1.82}                 & 48.83\scalebox{0.8}{$\pm$3.02}        & 92.00\scalebox{0.8}{$\pm$1.13}        & 55.10\scalebox{0.8}{$\pm$0.44}                 & 49.77\scalebox{0.8}{$\pm$0.76}   &63.58     \\
D-node     & 72.16\scalebox{0.8}{$\pm$0.28}                 & 48.33\scalebox{0.8}{$\pm$0.98}        & 90.25\scalebox{0.8}{$\pm$0.98}        & 53.26\scalebox{0.8}{$\pm$4.99}                 & 49.95\scalebox{0.8}{$\pm$1.70}   &62.79     \\
S-graph     & 68.50\scalebox{0.8}{$\pm$0.86}                 & 47.25\scalebox{0.8}{$\pm$3.78}        & 90.33\scalebox{0.8}{$\pm$0.87}        & 54.60\scalebox{0.8}{$\pm$3.15}                 & 49.67\scalebox{0.8}{$\pm$0.90}  &62.07      \\
M-mixup      & 70.83\scalebox{0.8}{$\pm$1.04}                 & 49.88\scalebox{0.8}{$\pm$1.34}        & 90.75\scalebox{0.8}{$\pm$1.78}        & 54.95\scalebox{0.8}{$\pm$0.86}                 & 49.81\scalebox{0.8}{$\pm$0.80} &63.24       \\
G-mixup      & \underline{71.94}\scalebox{0.8}{$\pm$3.00}                 & \textbf{50.46}\scalebox{0.8}{$\pm$1.49}        & \textbf{92.90}\scalebox{0.8}{$\pm$0.87}        & 55.49\scalebox{0.8}{$\pm$0.53}                 & \textbf{50.50}\scalebox{0.8}{$\pm$0.41}  &\underline{64.25}    \\
\midrule
RMA       & 71.17\scalebox{0.8}{$\pm$4.75}                 & \underline{49.89}\scalebox{0.8}{$\pm$2.50}        & 91.92\scalebox{0.8}{$\pm$1.70}        & 55.23\scalebox{0.8}{$\pm$0.58}       & 49.32\scalebox{0.8}{$\pm$1.45}     &63.51   \\
ABD          & 70.67\scalebox{0.8}{$\pm$3.25}                & 47.44\scalebox{0.8}{$\pm$1.84}       & 92.42\scalebox{0.8}{$\pm$0.14}       & 54.87\scalebox{0.8}{$\pm$0.45}          & 49.08\scalebox{0.8}{$\pm$0.86}   &62.90     \\
\scalebox{0.8}{ABD+SYN}      & 71.67\scalebox{0.8}{$\pm$3.25}           & 48.97\scalebox{0.8}{$\pm$2.97}       & 91.08\scalebox{0.8}{$\pm$1.23}        &\underline{55.90}\scalebox{0.8}{$\pm$0.95}   & 49.29\scalebox{0.8}{$\pm$1.00}    &63.38   \\
HMA & \textbf{72.62}\scalebox{0.8}{$\pm$0.48}          & 49.78\scalebox{0.8}{$\pm$0.74}       &  \underline{92.80}\scalebox{0.8}{$\pm$1.37} &\textbf{56.37}\scalebox{0.8}{$\pm$1.65} &  \underline{50.27}\scalebox{0.8}{$\pm$0.60}  &\textbf{64.37}\\

\bottomrule
\end{tabular}
}
\end{small}
\end{sc}
\end{center}
\vskip -0.1in
\label{GraphID}
\end{table*}

Furthermore, we follow G-mixup \cite{han2022g} and compare it with various augmentation methods, including DropEdge \cite{rong2019dropedge}, DropNode \cite{you2020graph}, Subgraph \cite{wang2020graphcrop}, and Manifold-Mixup \cite{wang2021mixup} on different datasets. 
All methods use the same GIN \cite{xu2018powerful} backbone. The results are summarized in \tablename~\ref{GraphID}. 
We can see the three components of HMA have a similar effect to CIFAR-10, and full HMA consistently outperforms vanilla GIN backbone.
By doing meaningful mixup with the help of elaborately designed graphon, G-mixup achieves better performance than simply mixing graphs in feature space \cite{wang2021mixup}. Although HMA also only operates in feature space, it significantly outperforms Manifold-Mixup and shows competitive results to G-Mixup.

\subsection{HMA Performs Competitively on OOD Tasks}
\label{ood-perform}
\vskip -0.1in
\begin{figure}[bht]
\begin{center}
\centerline{\includegraphics[width=0.9\columnwidth]{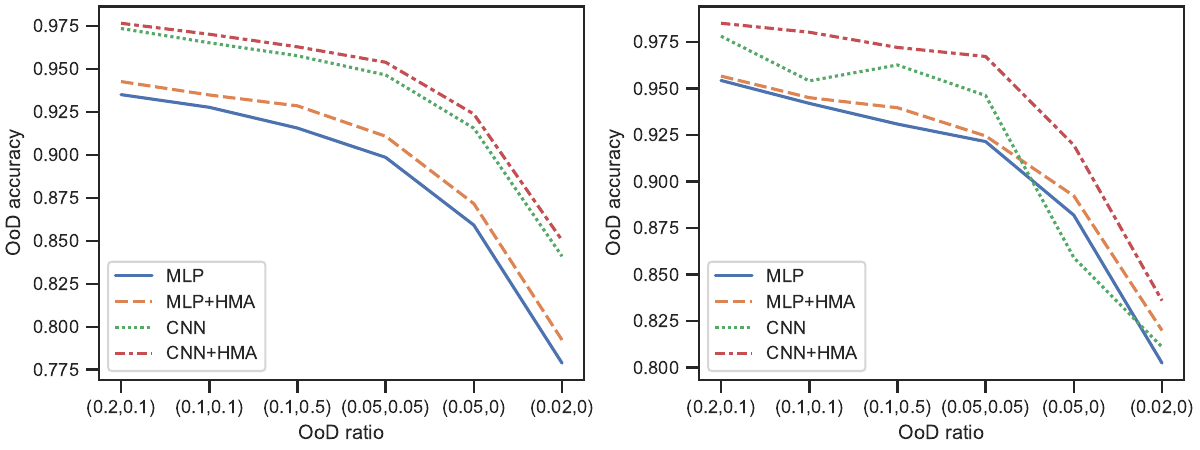}}
\vskip -0.1in
\caption{\ OOD results for colored MNIST (left: original size, right: downsized) experiments. We change the OOD sample ratio from ($0.2, 0.1$) to ($0, 0.02$)}
\label{CMNIST}
\end{center}
\vskip -0.3in
\end{figure}

We test the OOD performance of HMA on the Colored MNIST dataset introduced in IRM \cite{arjovsky2019invariant}. This task is distinct from traditional MNIST as it introduces misleading correlations between labels and colors. 
Specifically, digits 0-4 and 5-9 are considered two separate classes.
Two domains {$p_1, p_2$} can be accessed during training: in the first domain, green digits have a probability of $p_1$ to be in 0-4; in the second, the probability is $p_2$. 
During testing, there are two sets: 1. reversed color: green digits have a 90\% chance of being 0-4; 2. gray: blue and red images are turned into gray.
The original image size (28,28) is downsized for computational convenience to (14,14). 
HMA is tested with MLP from IRM \cite{arjovsky2019invariant} and CNN from \cite{gulrajani2020search} under both downsized and original image sizes.
We modify {$p_1, p_2$} from ($0.2, 0.1$) to ($0, 0.02$) to more accurately demonstrate the model's performance with distribution shifts.

This task provides a proof of concept for the semi-parametric approach in addressing out-of-distribution (OOD) tasks. While there is a distribution shift, there are still certain data points within the training set that share the same distribution as the test data. The semi-parametric approach effectively reduces the distribution shift by leveraging these data points as a support set.
The average accuracy of the two test domains is presented in \figurename~\ref{CMNIST}. With different backbones, image sizes, and distribution shifts, we can observe that HMA consistently improves OOD performance.

\begin{wraptable}{r}{8.5cm}
\vskip -0.1in
\caption{\ Acc of OOD algorithms on averaged degrees shift.}
\begin{center}
\begin{sc}
\begin{small}
\vskip -0.1in
\resizebox{0.5\columnwidth}{!}{
\begin{tabular}{lccc}
\toprule
Dataset   & SST5     & TWITTER     &avg\\
\midrule
ERM    & 43.72\scalebox{0.8}{$\pm$0.59} & \underline{63.54}\scalebox{0.8}{$\pm$2.10} & \underline{53.63}\\
ASAP   & \textbf{44.16}\scalebox{0.8}{$\pm$1.36} & 60.68\scalebox{0.8}{$\pm$2.10} & 52.42\\
DIR    & \underline{41.12}\scalebox{0.8}{$\pm$1.96} & 59.85\scalebox{0.8}{$\pm$2.98} & 50.49\\
\midrule
IRM    & 43.69\scalebox{0.8}{$\pm$1.26} & 63.50\scalebox{0.8}{$\pm$1.23} & 53.60\\
V-REX  & 43.28\scalebox{0.8}{$\pm$0.52} & 63.21\scalebox{0.8}{$\pm$1.57} & 53.25\\
IB-IRM & 40.85\scalebox{0.8}{$\pm$2.08} & 61.26\scalebox{0.8}{$\pm$1.20} & 51.06\\
CIGAv1 & 43.70\scalebox{0.8}{$\pm$1.98} & 62.02\scalebox{0.8}{$\pm$2.28} & 52.68\\
CIGAv2 & 43.30\scalebox{0.8}{$\pm$0.90} & 61.80\scalebox{0.8}{$\pm$2.03} & 52.55\\
\midrule
HMA   & 44.03\scalebox{0.8}{$\pm$1.01} & \textbf{65.12}\scalebox{0.8}{$\pm$2.17} & \textbf{54.58}\\
\bottomrule
\end{tabular}
}
\end{small}
\end{sc}
\end{center}
\vskip -0.2in
\label{OOD-SST}
\end{wraptable}

In the following graph tasks, we follow CIGA \cite{chenlearning} to examine HMA on diverse distribution shits.
Our baselines include ERM \cite{vapnik1999overview}, SOTA interpretable GNNs like ASAP Pooling\cite{ranjan2020asap}, and DIR\cite{wu2022discovering}, and SOTA OOD objectives like IRM\cite{arjovsky2019invariant}, v-Rex\cite{krueger2021out}, and IB-IRM\cite{ahuja2021invariance}.

In \tablename~\ref{OOD-mat}, we examine HMA on graph size shift scenarios on datasets converted from TU benchmarks\cite{morris2020tudataset}. 
Distribution shifts are generated through size-specific dataset splits following \cite{chenlearning} \cite{yehudai2021local}: the training set comprises graphs with sizes smaller than the 50th percentile, 10\% of which are held-out for the validation set, and the test set consists of those with sizes larger than the 90th percentile. Matthews correlation coefficient \cite{bevilacqua2021size} is adopted due to the class imbalance.
In \tablename~\ref{OOD-SST}, we further test HMA on larger real-world datasets with more complicated node degree biases. We use the Graph-SST datasets following CIGA\cite{chenlearning}. 
Graphs are assigned based on their average degree to generate distribution shifts: the training set includes graphs with average degrees smaller than the 50th percentile, the validation set includes those with average degrees larger than the 50th percentile and smaller than the 80th percentile, and the test set includes the left.
Graphs in Twitter are assigned in an inversed order to investigate OOD generalization ability from large degree graphs to small ones.

ERM is a strong baseline for these real-world tasks compared with specifically designed algorithms \cite{gulrajani2020search}. 
Our HMA consistently improves ERM baselines without introducing complex OOD objectives or learning strategies.

\begin{table}[h]
\caption{\ Matthews correlation coefficients of different OOD algorithms on graph size shifts for real-world graphs. We strictly follow CIGA \cite{chenlearning} and reproduce the results of ERM, GIGAv1\&v2.}
\vskip -0.2in
\begin{center}
\begin{sc}
\resizebox{0.7\columnwidth}{!}{
\begin{tabular}{lcccccc}
\toprule
Dataset  & NC11                      & NC109            & proteins         & DD             &avg  \\
\midrule
ERM       & 0.15\scalebox{0.8}{$\pm$0.03}                 & 0.16\scalebox{0.8}{$\pm$0.07}        & 0.19\scalebox{0.8}{$\pm$0.07}        & \underline{0.29}\scalebox{0.8}{$\pm$0.1}&  0.198\\
ASAP      & 0.16\scalebox{0.8}{$\pm$0.10}                & 0.15\scalebox{0.8}{$\pm$0.07}       & 0.22\scalebox{0.8}{$\pm$0.16}       & 0.21\scalebox{0.8}{$\pm$0.08} & 0.185\\
DIR     & \underline{0.21}\scalebox{0.8}{$\pm$0.06}       & 0.13\scalebox{0.8}{$\pm$0.05}       & 0.25\scalebox{0.8}{$\pm$0.14}       & 0.20\scalebox{0.8}{$\pm$0.10} & 0.198\\
\midrule
IRM       & 0.17\scalebox{0.8}{$\pm$0.02}                 & 0.14\scalebox{0.8}{$\pm$0.01}       & 0.21\scalebox{0.8}{$\pm$0.09}      & 0.22\scalebox{0.8}{$\pm$0.08} & 0.185\\
V-REX     & 0.15\scalebox{0.8}{$\pm$0.04}                & 0.15\scalebox{0.8}{$\pm$0.04}       & 0.22\scalebox{0.8}{$\pm$0.06}       & 0.21\scalebox{0.8}{$\pm$0.07} & 0.183\\
IB-IRM    & 0.12\scalebox{0.8}{$\pm$0.04}                & 0.15\scalebox{0.8}{$\pm$0.06}       & 0.21\scalebox{0.8}{$\pm$0.06}       & 0.15\scalebox{0.8}{$\pm$0.13} & 0.158\\
CIGAv1    & 0.25\scalebox{0.8}{$\pm$0.08}                 & \underline{0.26}\scalebox{0.8}{$\pm$0.09}        & \textbf{0.35}\scalebox{0.8}{$\pm$0.17}        & 0.21\scalebox{0.8}{$\pm$0.12} & \textbf{0.268}\\
CIGAv2    & \textbf{0.28}\scalebox{0.8}{$\pm$0.10}               & \textbf{0.27}\scalebox{0.8}{$\pm$0.10}       & 0.30\scalebox{0.8}{$\pm$0.09}        & 0.19\scalebox{0.8}{$\pm$0.12}  & \underline{0.260}\\
\midrule
HMA      & 0.19\scalebox{0.8}{$\pm$0.05}                 & 0.18\scalebox{0.8}{$\pm$0.04}        & \underline{0.34}\scalebox{0.8}{$\pm$0.07}        & \textbf{0.32}\scalebox{0.8}{$\pm$0.03} &0.258\\
\bottomrule
\end{tabular}}
\end{sc}
\end{center}
\vskip -0.1in
\label{OOD-mat}
\end{table}

\vskip -0.3in

\subsection{Heterogenous Memory Shows Reasonable Properties}
\label{memory properties}
\vskip -0.1in
We use T-SNE \cite{van2008visualizing} to visualize the relationships between the synthetic memory features and real data features in $C^2$. 
We excluded the label embeddings to focus on the synthetic memory slots since they would result in identical embeddings for instances with the same label and affect the T-SNE results.

\begin{figure}[h]
\begin{center}
\centerline{
\includegraphics[width=0.35\columnwidth]{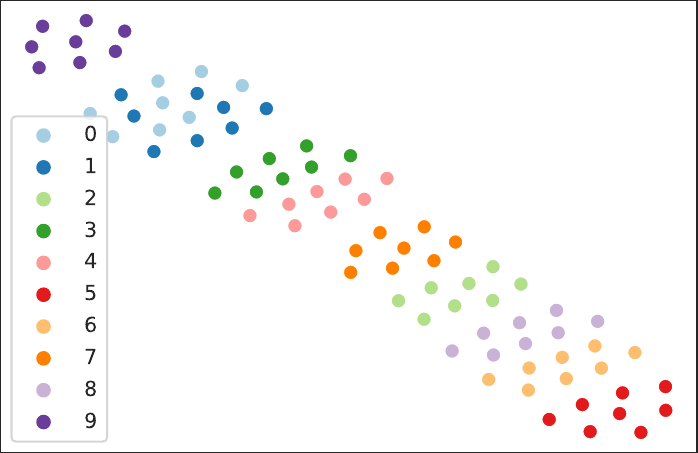}
\includegraphics[width=0.35\columnwidth]{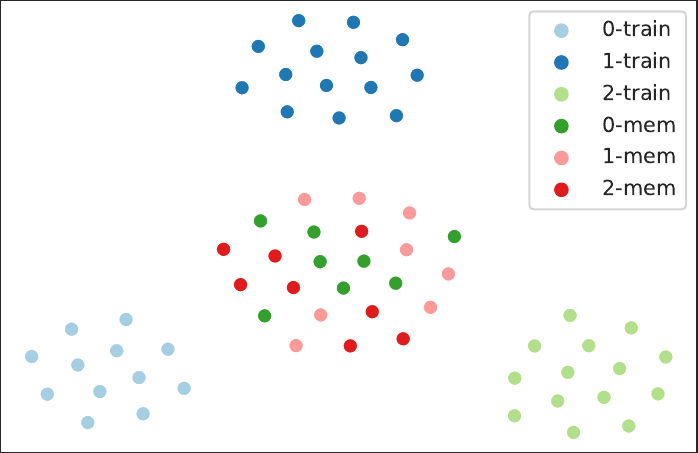}
}
\caption{\ T-SNE on CIFAR-10 for synthetic memory features (left) and real data features with synthetic memory features (right). '0-train' and '0-mem' refer to real and memory data features, respectively. The preceding label corresponds to the class from CIFAR10.}
\label{T-SNE_CIFAR}
\end{center}
\vskip -0.2in
\end{figure}

We present the synthetic memory slots' relationships in \figurename~\ref{T-SNE_CIFAR} (left). It can be observed that slots belonging to the same class cluster together, indicating that the synthetic memory slots can learn class-specific information with the help of label embeddings.
In \figurename~\ref{T-SNE_CIFAR} (right), we show the relationships between three classes of synthetic memory slots and their corresponding real data features.
It can be observed that different classes of real data features form separate clusters, while three separate clusters in \figurename~\ref{T-SNE_CIFAR} (left) merge into one. 
This suggests that the learned memory in the synthetic memory slots differs from the real data features. The improvement in \ref{general} and \ref{ood-perform} may come from the information beyond the instance level in the synthetic memory slots.

\begin{figure}
\begin{center}
\centerline{\includegraphics[width=0.35\columnwidth]{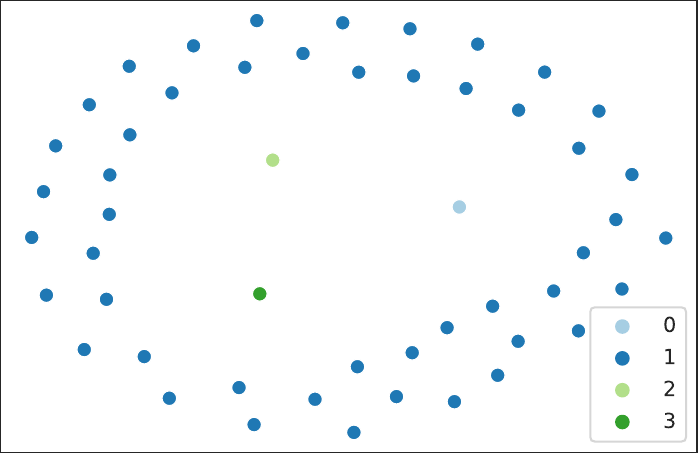}
 \includegraphics[width=0.35\columnwidth]{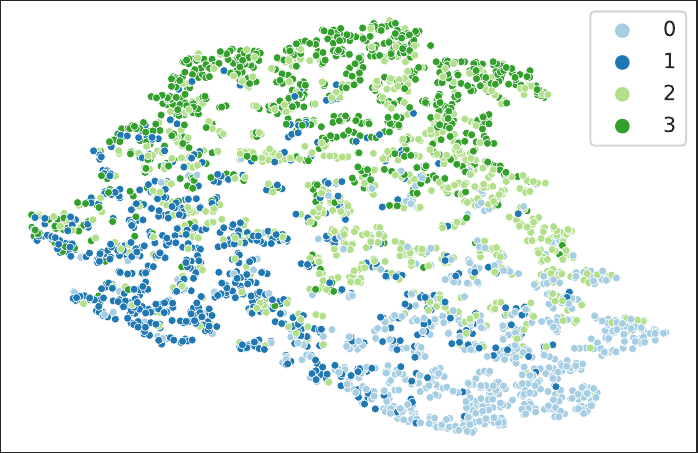}}
\caption{\ T-SNE on REDD-M5 for synthetic memory features (left) and input features (right)}
\label{T-SNE_graph}
\end{center}
\vskip -0.2in
\end{figure}

We also visualize the synthetic memory slots features learned in graph tasks in \figurename~\ref{T-SNE_graph}. Although the synthetic memory slots of class 1 scatter over other classes' clusters \figurename~\ref{T-SNE_graph} left, there is an interesting observation that the features of class 1 mix with the other three classes in the feature space correspondingly, as shown in \figurename~\ref{T-SNE_graph} right.
The correspondence indicates that the synthetic memory slots might capture class-level information, similar to prototype representations \cite{snell2017prototypical}.

\begin{figure*}
\begin{center}
\centerline{\includegraphics[width=\columnwidth]{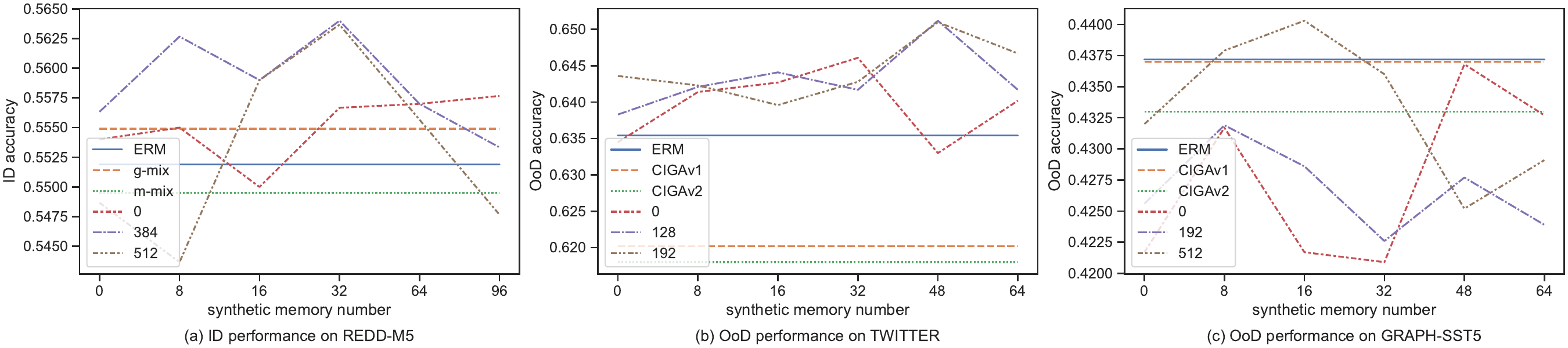}}
\caption{\ Performance of HMA on REDD-M5, Twitter and SST5 with different memory hyperparameters. The different colored dashed lines represent different sizes $m_1$ of real Memory Buffer (0 means no RMA), and it can be seen that introducing $m_1$ can make the model more stable; the horizontal axis represents different $m_2$ (0 represents using ABD without synthetic memory)}
\label{change}
\end{center}
\vskip -0.2in
\end{figure*}

\textbf{Ablation Study.} The in-distribution performances on REDD-M5 are depicted in \figurename~\ref{change}(a) for different memory configurations, while the out-of-distribution performances on Twitter and SST are illustrated in \figurename~\ref{change}(b) and (c) respectively.
It can be seen that while a suitable increase in the synthetic memory size $m_2$ can bring improvement, larger $m_2$ will hurt HMA's performance.
One potential explanation could be that a large $m_2$ increases the optimization difficulty of ABD.
Furthermore, it can be seen that with the addition of real memory augmentation, an increase in $m_2$ provides more improvement.

From \tablename~\ref{ViT-full} and \tablename~\ref{GraphID}, we can find that singly using ABD or SMA proves beneficial in certain scenarios but not significant. The possible reason for this observation is the sensitivity of semi-parametric approaches. By solely relying on SMA without any data filtering, the quality of the non-parametric elements remains unassured. However, upon incorporating RMA, the effectiveness of HMA surpasses that of individual deployments of RMA and SMA. This aligns with our conjecture that RMA explicitly offers more information to the synthetic memory learning process.

\section{Discussions, Conclusions, Limitations and Future work}
\vskip -0.1in
\textbf{Two-stage Augmentation.} While it seems more natural to perform both RMA and SMA in one layer, \textit{i.e.} to compute MHA with features aggregated from the input features, buffer features and the learnable features, the overall complexity is quadratic due to the SMA operations. 
It is possible to alternatively use linear approximations to standard attention (\textit{e.g.}, Performer \cite{49882}), it is beyond our scope and we leave it to future work.

We propose a universal memory augmentation method that can be seamlessly integrated with various backbone architectures. 
SMA leverages parametric and non-parametric approaches in the complementary combination of Attention Between Datapoints and learnable memory, thereby enabling the capture of class-specific and task-aware information, which is beneficial for ID and OOD downstream tasks.
Additionally, based on momentum memory queue and attention-based reading, RMA enhances SMA with cross-batch information.
Our experimental results demonstrate that HMA can consistently improve the performance of backbones and performs competitively with task-specific best-performing designs.

\bibliography{papers}

\begin{thebibliography}{10}

\bibitem{ahuja2021invariance}
Kartik Ahuja, Ethan Caballero, Dinghuai Zhang, Jean-Christophe Gagnon-Audet,
  Yoshua Bengio, Ioannis Mitliagkas, and Irina Rish.
\newblock Invariance principle meets information bottleneck for
  out-of-distribution generalization.
\newblock {\em Advances in Neural Information Processing Systems},
  34:3438--3450, 2021.

\bibitem{arbel2020generalized}
Michael Arbel, Liang Zhou, and Arthur Gretton.
\newblock Generalized energy based models.
\newblock {\em arXiv preprint arXiv:2003.05033}, 2020.

\bibitem{arjovsky2019invariant}
Martin Arjovsky, L{\'e}on Bottou, Ishaan Gulrajani, and David Lopez-Paz.
\newblock Invariant risk minimization.
\newblock {\em arXiv preprint arXiv:1907.02893}, 2019.

\bibitem{bevilacqua2021size}
Beatrice Bevilacqua, Yangze Zhou, and Bruno Ribeiro.
\newblock Size-invariant graph representations for graph classification
  extrapolations.
\newblock In {\em International Conference on Machine Learning}, pages
  837--851. PMLR, 2021.

\bibitem{bulatov2022recurrent}
Aydar Bulatov, Yury Kuratov, and Mikhail Burtsev.
\newblock Recurrent memory transformer.
\newblock {\em Advances in Neural Information Processing Systems},
  35:11079--11091, 2022.

\bibitem{burtsev2020memory}
Mikhail~S Burtsev, Yuri Kuratov, Anton Peganov, and Grigory~V Sapunov.
\newblock Memory transformer.
\newblock {\em arXiv preprint arXiv:2006.11527}, 2020.

\bibitem{cazenavette2022dataset}
George Cazenavette, Tongzhou Wang, Antonio Torralba, Alexei~A Efros, and
  Jun-Yan Zhu.
\newblock Dataset distillation by matching training trajectories.
\newblock In {\em Proceedings of the IEEE/CVF Conference on Computer Vision and
  Pattern Recognition}, pages 4750--4759, 2022.

\bibitem{chenlearning}
Yongqiang Chen, Yonggang Zhang, Yatao Bian, Han Yang, MA~KAILI, Binghui Xie,
  Tongliang Liu, Bo~Han, and James Cheng.
\newblock Learning causally invariant representations for out-of-distribution
  generalization on graphs.
\newblock In {\em Advances in Neural Information Processing Systems}.

\bibitem{49882}
Krzysztof~Marcin Choromanski, Valerii Likhosherstov, David~Martin Dohan,
  Xingyou Song, Andreea Gane, Tamas Sarlos, Peter Hawkins, Jared~Quincy Davis,
  Afroz Mohiuddin, Lukasz Kaiser, David Belanger, Lucy Colwell, and Adrian
  Weller.
\newblock Rethinking attention with performers.
\newblock In {\em accepted to ICLR 2021 (oral presentation)}, 2021.

\bibitem{cimpoi14describing}
M.~Cimpoi, S.~Maji, I.~Kokkinos, S.~Mohamed, , and A.~Vedaldi.
\newblock Describing textures in the wild.
\newblock In {\em Proceedings of the {IEEE} Conf. on Computer Vision and
  Pattern Recognition ({CVPR})}, 2014.

\bibitem{dai2019transformer}
Zihang Dai, Zhilin Yang, Yiming Yang, Jaime Carbonell, Quoc~V Le, and Ruslan
  Salakhutdinov.
\newblock Transformer-xl: Attentive language models beyond a fixed-length
  context.
\newblock {\em arXiv preprint arXiv:1901.02860}, 2019.

\bibitem{damianou2013deep}
Andreas Damianou and Neil~D Lawrence.
\newblock Deep gaussian processes.
\newblock In {\em Artificial intelligence and statistics}, pages 207--215.
  PMLR, 2013.

\bibitem{deng2009imagenet}
Jia Deng, Wei Dong, Richard Socher, Li-Jia Li, Kai Li, and Li~Fei-Fei.
\newblock Imagenet: A large-scale hierarchical image database.
\newblock In {\em 2009 IEEE conference on computer vision and pattern
  recognition}, pages 248--255. Ieee, 2009.

\bibitem{devlin2018bert}
Jacob Devlin, Ming-Wei Chang, Kenton Lee, and Kristina Toutanova.
\newblock Bert: Pre-training of deep bidirectional transformers for language
  understanding.
\newblock {\em arXiv preprint arXiv:1810.04805}, 2018.

\bibitem{dosovitskiy2020image}
Alexey Dosovitskiy, Lucas Beyer, Alexander Kolesnikov, Dirk Weissenborn,
  Xiaohua Zhai, Thomas Unterthiner, Mostafa Dehghani, Matthias Minderer, Georg
  Heigold, Sylvain Gelly, et~al.
\newblock An image is worth 16x16 words: Transformers for image recognition at
  scale.
\newblock {\em arXiv preprint arXiv:2010.11929}, 2020.

\bibitem{drozdov2022you}
Andrew Drozdov, Shufan Wang, Razieh Rahimi, Andrew McCallum, Hamed Zamani, and
  Mohit Iyyer.
\newblock You can't pick your neighbors, or can you? when and how to rely on
  retrieval in the $ k $ nn-lm.
\newblock {\em arXiv preprint arXiv:2210.15859}, 2022.

\bibitem{fan2021augmenting}
Angela Fan, Claire Gardent, Chlo{\'e} Braud, and Antoine Bordes.
\newblock Augmenting transformers with knn-based composite memory for dialog.
\newblock {\em Transactions of the Association for Computational Linguistics},
  9:82--99, 2021.

\bibitem{gilmer2017neural}
Justin Gilmer, Samuel~S Schoenholz, Patrick~F Riley, Oriol Vinyals, and
  George~E Dahl.
\newblock Neural message passing for quantum chemistry.
\newblock In {\em International conference on machine learning}, pages
  1263--1272. PMLR, 2017.

\bibitem{gulrajani2020search}
Ishaan Gulrajani and David Lopez-Paz.
\newblock In search of lost domain generalization.
\newblock In {\em International Conference on Learning Representations}, 2020.

\bibitem{gupta2020gmat}
Ankit Gupta and Jonathan Berant.
\newblock Gmat: Global memory augmentation for transformers.
\newblock {\em arXiv preprint arXiv:2006.03274}, 2020.

\bibitem{han2022g}
Xiaotian Han, Zhimeng Jiang, Ninghao Liu, and Xia Hu.
\newblock G-mixup: Graph data augmentation for graph classification.
\newblock {\em arXiv preprint arXiv:2202.07179}, 2022.

\bibitem{he2020momentum}
Kaiming He, Haoqi Fan, Yuxin Wu, Saining Xie, and Ross Girshick.
\newblock Momentum contrast for unsupervised visual representation learning.
\newblock In {\em Proceedings of the IEEE/CVF conference on computer vision and
  pattern recognition}, pages 9729--9738, 2020.

\bibitem{he2016deep}
Kaiming He, Xiangyu Zhang, Shaoqing Ren, and Jian Sun.
\newblock Deep residual learning for image recognition.
\newblock In {\em Proceedings of the IEEE conference on computer vision and
  pattern recognition}, pages 770--778, 2016.

\bibitem{hochreiter1997long}
Sepp Hochreiter and J{\"u}rgen Schmidhuber.
\newblock Long short-term memory.
\newblock {\em Neural computation}, 9(8):1735--1780, 1997.

\bibitem{hou2022batchformer}
Zhi Hou, Baosheng Yu, and Dacheng Tao.
\newblock Batchformer: Learning to explore sample relationships for robust
  representation learning.
\newblock In {\em Proceedings of the IEEE/CVF Conference on Computer Vision and
  Pattern Recognition}, pages 7256--7266, 2022.

\bibitem{hu2020open}
Weihua Hu, Matthias Fey, Marinka Zitnik, Yuxiao Dong, Hongyu Ren, Bowen Liu,
  Michele Catasta, and Jure Leskovec.
\newblock Open graph benchmark: Datasets for machine learning on graphs.
\newblock {\em Advances in neural information processing systems},
  33:22118--22133, 2020.

\bibitem{humphreys2022large}
Peter Humphreys, Arthur Guez, Olivier Tieleman, Laurent Sifre, Th{\'e}ophane
  Weber, and Timothy Lillicrap.
\newblock Large-scale retrieval for reinforcement learning.
\newblock {\em Advances in Neural Information Processing Systems},
  35:20092--20104, 2022.

\bibitem{karpukhin2020dense}
Vladimir Karpukhin, Barlas O{\u{g}}uz, Sewon Min, Patrick Lewis, Ledell Wu,
  Sergey Edunov, Danqi Chen, and Wen-tau Yih.
\newblock Dense passage retrieval for open-domain question answering.
\newblock {\em arXiv preprint arXiv:2004.04906}, 2020.

\bibitem{khandelwal2019generalization}
Urvashi Khandelwal, Omer Levy, Dan Jurafsky, Luke Zettlemoyer, and Mike Lewis.
\newblock Generalization through memorization: Nearest neighbor language
  models.
\newblock {\em arXiv preprint arXiv:1911.00172}, 2019.

\bibitem{kipf2016semi}
Thomas~N Kipf and Max Welling.
\newblock Semi-supervised classification with graph convolutional networks.
\newblock {\em arXiv preprint arXiv:1609.02907}, 2016.

\bibitem{kossen2021self}
Jannik Kossen, Neil Band, Clare Lyle, Aidan~N Gomez, Thomas Rainforth, and
  Yarin Gal.
\newblock Self-attention between datapoints: Going beyond individual
  input-output pairs in deep learning.
\newblock {\em Advances in Neural Information Processing Systems},
  34:28742--28756, 2021.

\bibitem{krizhevsky2009learning}
Alex Krizhevsky, Geoffrey Hinton, et~al.
\newblock Learning multiple layers of features from tiny images.
\newblock 2009.

\bibitem{krueger2021out}
David Krueger, Ethan Caballero, Joern-Henrik Jacobsen, Amy Zhang, Jonathan
  Binas, Dinghuai Zhang, Remi Le~Priol, and Aaron Courville.
\newblock Out-of-distribution generalization via risk extrapolation (rex).
\newblock In {\em International Conference on Machine Learning}, pages
  5815--5826. PMLR, 2021.

\bibitem{lanchantin2021general}
Jack Lanchantin, Tianlu Wang, Vicente Ordonez, and Yanjun Qi.
\newblock General multi-label image classification with transformers.
\newblock In {\em Proceedings of the IEEE/CVF Conference on Computer Vision and
  Pattern Recognition}, pages 16478--16488, 2021.

\bibitem{le2019neural}
Hung Le, Truyen Tran, and Svetha Venkatesh.
\newblock Neural stored-program memory.
\newblock {\em arXiv preprint arXiv:1906.08862}, 2019.

\bibitem{li2017learning}
Junying Li, Deng Cai, and Xiaofei He.
\newblock Learning graph-level representation for drug discovery.
\newblock {\em arXiv preprint arXiv:1709.03741}, 2017.

\bibitem{lin2021bertgcn}
Yuxiao Lin, Yuxian Meng, Xiaofei Sun, Qinghong Han, Kun Kuang, Jiwei Li, and
  Fei Wu.
\newblock Bertgcn: Transductive text classification by combining gcn and bert.
\newblock {\em arXiv preprint arXiv:2105.05727}, 2021.

\bibitem{liu2021p}
Xiao Liu, Kaixuan Ji, Yicheng Fu, Weng~Lam Tam, Zhengxiao Du, Zhilin Yang, and
  Jie Tang.
\newblock P-tuning v2: Prompt tuning can be comparable to fine-tuning
  universally across scales and tasks.
\newblock {\em arXiv preprint arXiv:2110.07602}, 2021.

\bibitem{liu2021gpt}
Xiao Liu, Yanan Zheng, Zhengxiao Du, Ming Ding, Yujie Qian, Zhilin Yang, and
  Jie Tang.
\newblock Gpt understands, too.
\newblock {\em arXiv preprint arXiv:2103.10385}, 2021.

\bibitem{meng2021gnn}
Yuxian Meng, Shi Zong, Xiaoya Li, Xiaofei Sun, Tianwei Zhang, Fei Wu, and Jiwei
  Li.
\newblock Gnn-lm: Language modeling based on global contexts via gnn.
\newblock {\em arXiv preprint arXiv:2110.08743}, 2021.

\bibitem{morris2020tudataset}
Christopher Morris, Nils~M Kriege, Franka Bause, Kristian Kersting, Petra
  Mutzel, and Marion Neumann.
\newblock Tudataset: A collection of benchmark datasets for learning with
  graphs.
\newblock {\em arXiv preprint arXiv:2007.08663}, 2020.

\bibitem{munkhdalai2017meta}
Tsendsuren Munkhdalai and Hong Yu.
\newblock Meta networks.
\newblock In {\em International Conference on Machine Learning}, pages
  2554--2563. PMLR, 2017.

\bibitem{Nilsback08}
Maria-Elena Nilsback and Andrew Zisserman.
\newblock Automated flower classification over a large number of classes.
\newblock In {\em Indian Conference on Computer Vision, Graphics and Image
  Processing}, Dec 2008.

\bibitem{orhan2018simple}
Emin Orhan.
\newblock A simple cache model for image recognition.
\newblock {\em Advances in Neural Information Processing Systems}, 31, 2018.

\bibitem{pahde2021multimodal}
Frederik Pahde, Mihai Puscas, Tassilo Klein, and Moin Nabi.
\newblock Multimodal prototypical networks for few-shot learning.
\newblock In {\em Proceedings of the IEEE/CVF Winter Conference on Applications
  of Computer Vision}, pages 2644--2653, 2021.

\bibitem{parkhi12a}
Omkar~M. Parkhi, Andrea Vedaldi, Andrew Zisserman, and C.~V. Jawahar.
\newblock Cats and dogs.
\newblock In {\em IEEE Conference on Computer Vision and Pattern Recognition},
  2012.

\bibitem{peng2023semiparametric}
Guangyue Peng, Tao Ge, Si-Qing Chen, Furu Wei, and Houfeng Wang.
\newblock Semiparametric language models are scalable continual learners.
\newblock {\em arXiv preprint arXiv:2303.01421}, 2023.

\bibitem{qu2020coda}
Yanru Qu, Dinghan Shen, Yelong Shen, Sandra Sajeev, Jiawei Han, and Weizhu
  Chen.
\newblock Coda: Contrast-enhanced and diversity-promoting data augmentation for
  natural language understanding.
\newblock {\em arXiv preprint arXiv:2010.08670}, 2020.

\bibitem{ranjan2020asap}
Ekagra Ranjan, Soumya Sanyal, and Partha Talukdar.
\newblock Asap: Adaptive structure aware pooling for learning hierarchical
  graph representations.
\newblock In {\em Proceedings of the AAAI Conference on Artificial
  Intelligence}, volume~34, pages 5470--5477, 2020.

\bibitem{rastogi2022semi}
Richa Rastogi, Yuntian Deng, Ian Lee, Mert~R Sabuncu, and Volodymyr Kuleshov.
\newblock Semi-parametric deep neural networks in linear time and memory.
\newblock {\em arXiv preprint arXiv:2205.11718}, 2022.

\bibitem{rong2019dropedge}
Yu~Rong, Wenbing Huang, Tingyang Xu, and Junzhou Huang.
\newblock Dropedge: Towards deep graph convolutional networks on node
  classification.
\newblock In {\em International Conference on Learning Representations}, 2019.

\bibitem{sandler2022fine}
Mark Sandler, Andrey Zhmoginov, Max Vladymyrov, and Andrew Jackson.
\newblock Fine-tuning image transformers using learnable memory.
\newblock In {\em Proceedings of the IEEE/CVF Conference on Computer Vision and
  Pattern Recognition}, pages 12155--12164, 2022.

\bibitem{snell2017prototypical}
Jake Snell, Kevin Swersky, and Richard Zemel.
\newblock Prototypical networks for few-shot learning.
\newblock {\em Advances in neural information processing systems}, 30, 2017.

\bibitem{sprechmann2018memory}
Pablo Sprechmann, Siddhant~M Jayakumar, Jack~W Rae, Alexander Pritzel,
  Adria~Puigdomenech Badia, Benigno Uria, Oriol Vinyals, Demis Hassabis, Razvan
  Pascanu, and Charles Blundell.
\newblock Memory-based parameter adaptation.
\newblock {\em arXiv preprint arXiv:1802.10542}, 2018.

\bibitem{sun2019hierarchical}
Shengli Sun, Qingfeng Sun, Kevin Zhou, and Tengchao Lv.
\newblock Hierarchical attention prototypical networks for few-shot text
  classification.
\newblock In {\em Proceedings of the 2019 conference on empirical methods in
  natural language processing and the 9th international joint conference on
  natural language processing (EMNLP-IJCNLP)}, pages 476--485, 2019.

\bibitem{van2008visualizing}
Laurens Van~der Maaten and Geoffrey Hinton.
\newblock Visualizing data using t-sne.
\newblock {\em Journal of machine learning research}, 9(11), 2008.

\bibitem{vapnik1999overview}
Vladimir~N Vapnik.
\newblock An overview of statistical learning theory.
\newblock {\em IEEE transactions on neural networks}, 10(5):988--999, 1999.

\bibitem{vaswani2017attention}
Ashish Vaswani, Noam Shazeer, Niki Parmar, Jakob Uszkoreit, Llion Jones,
  Aidan~N Gomez, {\L}ukasz Kaiser, and Illia Polosukhin.
\newblock Attention is all you need.
\newblock {\em Advances in neural information processing systems}, 30, 2017.

\bibitem{Wang2022RetrievalenhancedGN}
Dingmin Wang, Shengchao Liu, Hanchen Wang, Bernardo~Cuenca Grau, Linfeng Song,
  Jian Tang, Song Le, and Qi~Li.
\newblock Retrieval-enhanced graph neural networks for graph property
  prediction.
\newblock 2022.

\bibitem{wang2018dataset}
Tongzhou Wang, Jun-Yan Zhu, Antonio Torralba, and Alexei~A Efros.
\newblock Dataset distillation.
\newblock {\em arXiv preprint arXiv:1811.10959}, 2018.

\bibitem{wang2020graphcrop}
Yiwei Wang, Wei Wang, Yuxuan Liang, Yujun Cai, and Bryan Hooi.
\newblock Graphcrop: Subgraph cropping for graph classification.
\newblock {\em arXiv preprint arXiv:2009.10564}, 2020.

\bibitem{wang2021mixup}
Yiwei Wang, Wei Wang, Yuxuan Liang, Yujun Cai, and Bryan Hooi.
\newblock Mixup for node and graph classification.
\newblock In {\em Proceedings of the Web Conference 2021}, pages 3663--3674,
  2021.

\bibitem{wang2022dualprompt}
Zifeng Wang, Zizhao Zhang, Sayna Ebrahimi, Ruoxi Sun, Han Zhang, Chen-Yu Lee,
  Xiaoqi Ren, Guolong Su, Vincent Perot, Jennifer Dy, et~al.
\newblock Dualprompt: Complementary prompting for rehearsal-free continual
  learning.
\newblock In {\em Computer Vision--ECCV 2022: 17th European Conference, Tel
  Aviv, Israel, October 23--27, 2022, Proceedings, Part XXVI}, pages 631--648.
  Springer, 2022.

\bibitem{wang2022learning}
Zifeng Wang, Zizhao Zhang, Chen-Yu Lee, Han Zhang, Ruoxi Sun, Xiaoqi Ren,
  Guolong Su, Vincent Perot, Jennifer Dy, and Tomas Pfister.
\newblock Learning to prompt for continual learning.
\newblock In {\em Proceedings of the IEEE/CVF Conference on Computer Vision and
  Pattern Recognition}, pages 139--149, 2022.

\bibitem{wu2020memformer}
Qingyang Wu, Zhenzhong Lan, Jing Gu, and Zhou Yu.
\newblock Memformer: The memory-augmented transformer.
\newblock {\em arXiv preprint arXiv:2010.06891}, 2020.

\bibitem{wu2022discovering}
Ying-Xin Wu, Xiang Wang, An~Zhang, Xiangnan He, and Tat-Seng Chua.
\newblock Discovering invariant rationales for graph neural networks.
\newblock {\em arXiv preprint arXiv:2201.12872}, 2022.

\bibitem{wu2022memorizing}
Yuhuai Wu, Markus~N Rabe, DeLesley Hutchins, and Christian Szegedy.
\newblock Memorizing transformers.
\newblock {\em arXiv preprint arXiv:2203.08913}, 2022.

\bibitem{wu2022efficient}
Yuxiang Wu, Yu~Zhao, Baotian Hu, Pasquale Minervini, Pontus Stenetorp, and
  Sebastian Riedel.
\newblock An efficient memory-augmented transformer for knowledge-intensive nlp
  tasks.
\newblock {\em arXiv preprint arXiv:2210.16773}, 2022.

\bibitem{wu2018unsupervised}
Zhirong Wu, Yuanjun Xiong, Stella~X Yu, and Dahua Lin.
\newblock Unsupervised feature learning via non-parametric instance
  discrimination.
\newblock In {\em Proceedings of the IEEE conference on computer vision and
  pattern recognition}, pages 3733--3742, 2018.

\bibitem{wu2021learning}
Zhonghua Wu, Xiangxi Shi, Guosheng Lin, and Jianfei Cai.
\newblock Learning meta-class memory for few-shot semantic segmentation.
\newblock In {\em Proceedings of the IEEE/CVF International Conference on
  Computer Vision}, pages 517--526, 2021.

\bibitem{xu2022boosting}
Jitao Xu, Josep~M Crego, and Jean Senellart.
\newblock Boosting neural machine translation with similar translations.
\newblock In {\em Proceedings of the 15th Biennial Conference of the
  Association for Machine Translation in the Americas (Volume 2: Users and
  Providers Track and Government Track)}, pages 282--292, 2022.

\bibitem{xu2018powerful}
Keyulu Xu, Weihua Hu, Jure Leskovec, and Stefanie Jegelka.
\newblock How powerful are graph neural networks?
\newblock In {\em International Conference on Learning Representations}, 2018.

\bibitem{yehudai2021local}
Gilad Yehudai, Ethan Fetaya, Eli Meirom, Gal Chechik, and Haggai Maron.
\newblock From local structures to size generalization in graph neural
  networks.
\newblock In {\em International Conference on Machine Learning}, pages
  11975--11986. PMLR, 2021.

\bibitem{you2020graph}
Yuning You, Tianlong Chen, Yongduo Sui, Ting Chen, Zhangyang Wang, and Yang
  Shen.
\newblock Graph contrastive learning with augmentations.
\newblock {\em Advances in Neural Information Processing Systems},
  33:5812--5823, 2020.

\bibitem{zaheer2020big}
Manzil Zaheer, Guru Guruganesh, Kumar~Avinava Dubey, Joshua Ainslie, Chris
  Alberti, Santiago Ontanon, Philip Pham, Anirudh Ravula, Qifan Wang, Li~Yang,
  et~al.
\newblock Big bird: Transformers for longer sequences.
\newblock {\em Advances in Neural Information Processing Systems},
  33:17283--17297, 2020.

\bibitem{zhao2021datasetb}
Bo~Zhao and Hakan Bilen.
\newblock Dataset condensation with differentiable siamese augmentation.
\newblock In {\em International Conference on Machine Learning}, pages
  12674--12685. PMLR, 2021.

\bibitem{zhao2023dataset}
Bo~Zhao and Hakan Bilen.
\newblock Dataset condensation with distribution matching.
\newblock In {\em Proceedings of the IEEE/CVF Winter Conference on Applications
  of Computer Vision}, pages 6514--6523, 2023.

\bibitem{zhao2021dataset}
Bo~Zhao, Konda~Reddy Mopuri, and Hakan Bilen.
\newblock Dataset condensation with gradient matching.
\newblock {\em ICLR}, 1(2):3, 2021.

\end{thebibliography}
\bibliographystyle{plain}

\section{Appendix}

\subsection{Discussions, Limitations and Future works}

Our work also shares similarities with the approach in \cite{lanchantin2021general} that both incorporate labels into the learning process and utilize dependencies within the data. However, \cite{lanchantin2021general} explores the presence of different class labels within the same image, for example, ’dolphin is unlikely to co-occur with grass, while a knife is more likely to appear next to a fork‘, while HMA upon the observation in NPT and consider the correlations between different data points to improve performance.

We have observed that HMA brings more significant improvements when the dataset size is small. This may be due to the current design of HMA, which limits the capacity of the non-parametric components or struggles to capture complex information in larger datasets. In some scenarios, HMA's performance is still weaker than the SOTA methods. This could be attributed to the "trade-off" between efficiency and effectiveness in the two stages of HMA's design: in SMA, we introduce learnable memory tokens to approximate dataset-related information, while in RMA, we use a momentum memory buffer to approximate cross-batch information. These design choices allow HMA to provide improvements when combined with various backbones without significant computational overhead.

Future work could focus on more finely constructing the non-parametric components. For example, using Dataset Distillation \cite{wang2018dataset} to initialize learnable memory, introducing a retriever \cite{peng2023semiparametric} to enhance the quality of the non-parametric components, incorporating contrastive loss to assist in non-parametric learning, and obtaining memory with improved interpretability. Furthermore, introducing learnable non-parametric components like HMA can also be applied to scenarios and used to improve the efficiency of semi-parametric methods like retrievers \cite{peng2023semiparametric}.

\subsection{Related Work – Continued}

\subsubsection{Neural Memory}
Memory mechanism is widely used due to its capacity to process long-dependency information.
LSTM \cite{hochreiter1997long}, Transformer XL \cite{dai2019transformer}, Recurrent Memory Transformer \cite{bulatov2022recurrent} enable the model to handle long context information by passing hidden states.
\cite{zaheer2020big, gupta2020gmat, burtsev2020memory} incorporate global memory tokens to reduce the computational cost of full self-attention and synthesize information within the same window.
Memformer \cite{wu2020memformer},  Memorizing Transformer \cite{wu2022memorizing} introduces external memory to mitigate the issues caused by the limited transformer window size.
Other works \cite{wu2022efficient, fan2021augmenting} use the preprocessed knowledge base as additional memory, enabling the model to access more knowledge in dialogue and question-answering tasks, analogous to retrieval methods.

While there are many similarities between the synthetic memory in HMA and other learnable memories or soft prompts, our method exhibits two key differences: 1. We introduce learnable synthetic memory slots specifically in the semi-parametric ABD process to capture \emph{dataset-relevant} information effectively. 2. The synthetic memory learned during the ABD process aids the semi-parametric approach without relying on large batch sizes or additional selection methods. Furthermore, experimental results indicate that with the indirect guidance of label embeddings, the learned synthetic memory in HMA also contains class-specific information.

\subsubsection{Dataset Distillation}
Dataset distillation (DS) attempts to condense the information of an entire dataset into a few synthetic images. 
Initial work \cite{wang2018dataset} treated the synthetic images as hyperparameters and used gradient-based hyperparameter optimization methods to update them. 
Subsequent studies \cite{zhao2021dataset, zhao2021datasetb, cazenavette2022dataset} shift the optimization goal to minimizing the training gradient differences. 
However, these methods have two drawbacks: they are expensive due to the extra networks and second-order gradients, and the obtained synthetic images are only focused on aiding network training, which could be seen as a \emph{condensed gradient} with no direct connection to downstream tasks. 
\cite{wang2018dataset,zhao2023dataset} optimized the synthetic images by reducing the distribution distance in the feature space.
Drawing inspiration from DS, we use class-specific synthetic memory slots in ABD to capture information beyond individual data points. Moreover, our synthetic memory slots are optimized with gradients directly from the current task.

\subsection{Experiments – Continued}

\subsubsection{Training Details}

The training process for all experiments remains consistent with the original experimental settings, allowing HMA to be directly integrated with various backbones.

In the ViT tuning experiments, the baselines follow the original ViT's \cite{dosovitskiy2020image} training step settings (CIFAR100=10000, pets37=500, flowers102=500, dtd=1000). The learning rate was searched within the range of [0.001, 0.003, 0.01, 0.03]. In the first few HMA experiments, we found that lr=0.003 in full-tuning experiments and lr=0.01 in part-tuning experiments yielded satisfactory results. Therefore, for all HMA experiments, only these two learning rates were used respectively to reduce hyperparameter searches. For the size of the real memory buffer $m_1$, we searched within the range of the original experiment's default batch size multiplied by [1, 2, 3]. For the number of synthetic memories $m_2$, we searched within [4, 8, 16, 32]. For the label embedding size $d_2$, we used $d_2=64$ in ViT and all subsequent experiments.

In the remaining experiments, all hyperparameters were kept the same as the original settings, and the two hyperparameters introduced by HMA, $m_1$ and $m_2$, were searched similarly as in ViT. Specifically, in the graph experiments, where the number of label categories is smaller, we adjusted the search range of $m_2$ to [8, 16, 32, 64].

\subsubsection{Computation}

Experiments were conducted on NVIDIA Tesla P100 16 GB GPUs (one experiment was conducted on one GPU).

In the tuning experiment of ViT \cite{dosovitskiy2020image} on the DTD \cite{cimpoi14describing}  dataset, the baseline, which involves tuning the head and the CLS token, has 87.5 M parameters. Among them, there are 36.9 K trainable parameters. These whole occupy approximately 4539 MiB of GPU memory. The training process for 1000 steps takes around 360 seconds. After incorporating HMA, the total number of parameters increases to 119 M, with 32.4 M trainable parameters. It occupies approximately 5159 MiB of GPU memory, and the training time for 1000 steps is comparable to the backbone alone without any significant difference. Although it introduces additional parameters, our ablation study demonstrates that directly increasing the number of parameters may lead to worse performance. Additionally, the additional memory overhead caused by HMA is relatively small.

It is worth noting that the number of parameters introduced by HMA depends only on the output feature dimension of the backbone and the label embedding. If the feature dimension is reduced, the number of parameters and computational overhead of HMA can be further reduced.

In the graph OOD experiment on the Graph-Twitter dataset, the standard ERM \cite{vapnik1999overview} method takes approximately 10 minutes for a single experiment. The state-of-the-art method, CIGA \cite{chenlearning}, which involves additional training stages and loss takes around 50 minutes for a single experiment. On the other hand, HMA completes a single experiment in approximately 20 minutes. Despite taking significantly less time than CIGA, HMA achieves superior performance on this dataset. Similarly, for the Graph-SST5 dataset, the single experiment runtime for ERM, CIGA, and HMA is approximately 15 minutes, 140 minutes, and 60 minutes, respectively.

\subsubsection{Extended Visualization Results}

\begin{figure}[h]
\begin{center}
\centerline{
\includegraphics[width=0.4\columnwidth]{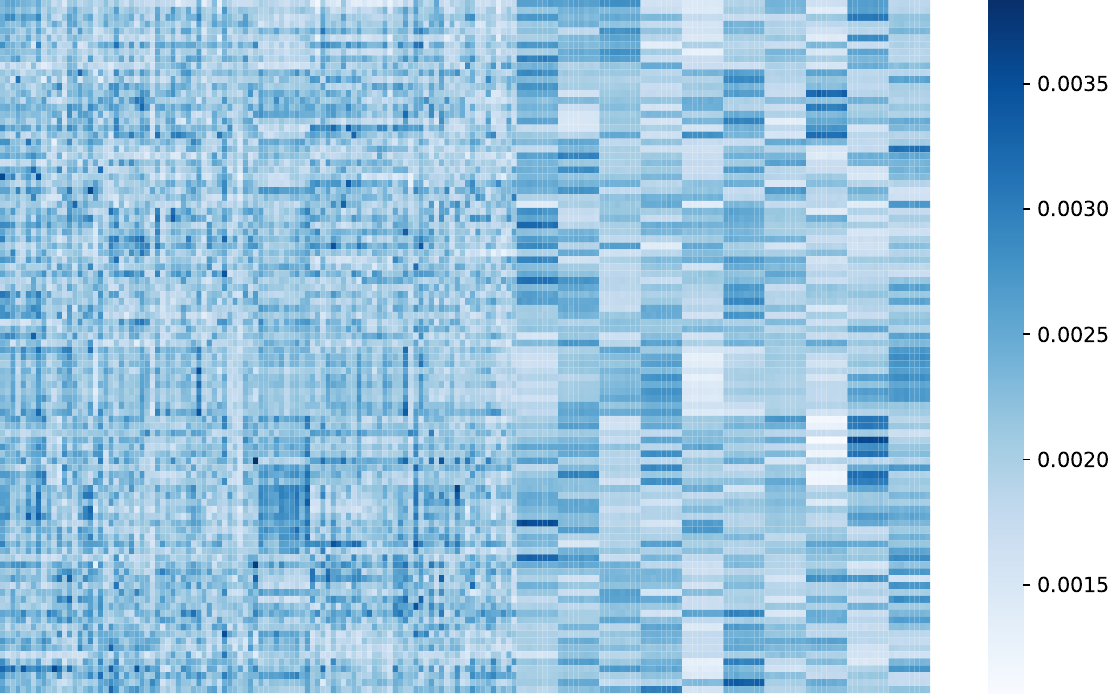}
\includegraphics[width=0.4\columnwidth]{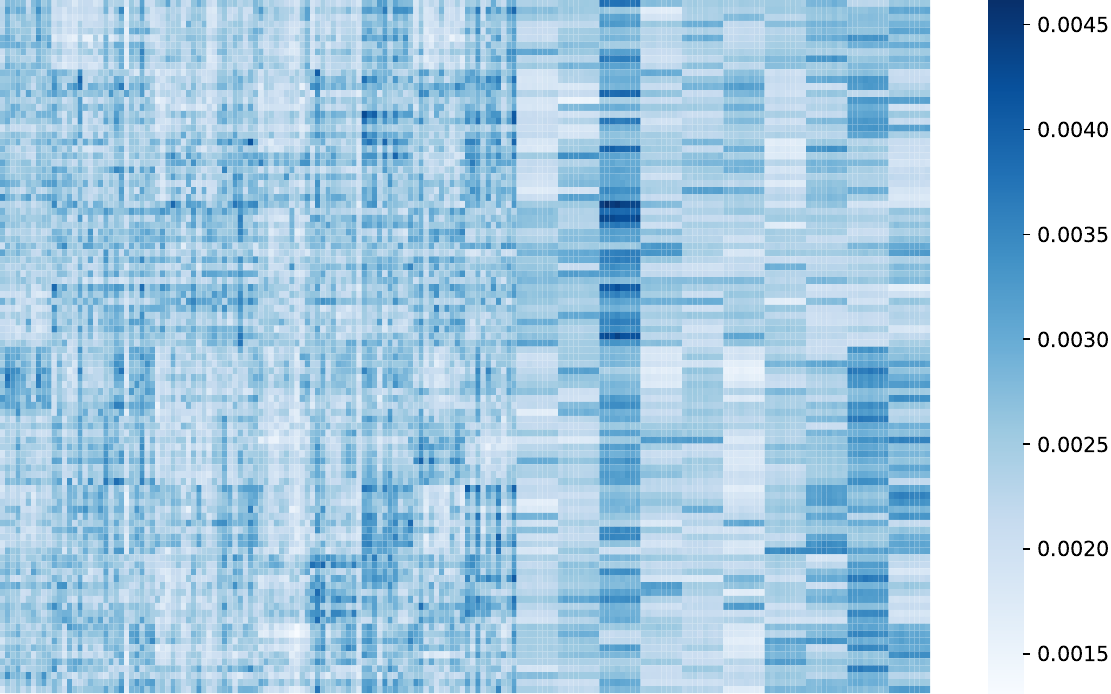}
}
\caption{\ The visualization of attention scores: the left part illustrates the results for the top 10 classes with 10 samples each on the DTD dataset, while the right part depicts the visualization for the top 10 classes with 10 samples each on the Pets37 dataset. The y-axis corresponds to the input samples, with samples from each class grouped. The x-axis represents the corresponding samples and synthetic memory. The left side of each image exhibits the attention scores among the samples, while the right side illustrates the scores between the samples and synthetic memory.}
\label{Attention-vis}
\end{center}
\vskip -0.2in
\end{figure}

We also visualized the attention scores in SMA, as shown in Figure~\ref{Attention-vis}. We selected 10 images from 10 different classes in two datasets, forming a 10x10 batch for classification, and visualized the attention scores in SMA. We can observe the following: (1) There is a clear distinction between the attention patterns among real data and the patterns between real data and synthetic memory (left vs. right). (2) Data from the same class exhibit more similar attention scores towards the synthetic memory (as evident in the attention scores on the right, with noticeable block patterns along the vertical axis). Observation (1) may be attributed to the fact that synthetic memory of the same class shares the same label embedding, and similar initialization methods lead to the learning of similar memory features. As a result, keys generated from the same class of synthetic memories tend to exhibit similar behaviors. Observation (2) suggests that synthetic memory captures information at the class level. Additionally, the ability of SMA to help inputs find similar reference sample is likely a contributing factor to the improvement in OOD performance.

\subsubsection{Extended Results}

\begin{table}[h]
  \caption{Accuracy of Resnet18 and HMA on CIFAR-10.There are three components of our approach: Real Memory Augmentation (RMA), Attention between Data-points (ABD), and Synthetic Memory (SYN).}
  \label{CIFARAcc}
\begin{center}
\begin{small}
\begin{sc}
\begin{tabular}{cccc}
\toprule
RMA  & ABD      & SYN  &   ACC      \\
\midrule
           &          &          & 95.43\scalebox{0.8}{$\pm$0.01} \\
           & $\surd$  &          & 95.40\scalebox{0.8}{$\pm$0.15} \\
$\surd$    & $\surd$  &          & 95.43\scalebox{0.8}{$\pm$0.14} \\
$\surd$    &          &          & 95.35\scalebox{0.8}{$\pm$0.03} \\
           & $\surd$  & $\surd$  & 95.49\scalebox{0.8}{$\pm$0.17} \\
$\surd$    & $\surd$  & $\surd$  & \textbf{95.54}\scalebox{0.8}{$\pm$0.11}\\
\bottomrule
\end{tabular}
\end{sc}
\end{small}
\end{center}
\end{table}

On CIFAR-10, we adopted a ResNet-18 encoder, similar to the one used in NPT\cite{kossen2021self}. 
From the accuracy results in \tablename~\ref{CIFARAcc}, we can observe the impact of the three components of our approach: Real Memory Augmentation (RMA), Attention between Datapoints (ABD), and Synthetic Memory (SYN).
If none of them is used, the model degrades to the backbone. 
We have the following observations: using ABD alone can hurt the performance similar to NPT; adding SMA during ABD can bring an improvement; while RMA alone doesn't bring a significant effect, combining RMA and SMA yields the best results. Although the variance is larger than the backbone, our synthetic memory does improve the best results.

\begin{table*}
\caption{Accuarcy of tuning head and cls token of pretrained ViT on 4 image datasets.}
\begin{center}
\begin{small}
\begin{sc}
\resizebox{0.8\columnwidth}{!}{
\begin{tabular}{lcccccc}
\toprule
dataset     & dtd                     & pets37            & flowers102            & cifar100         &avg     \\
\midrule
vit       &74.01\scalebox{0.8}{$\pm$0.29}       &89.17\scalebox{0.8}{$\pm$0.07}     &98.53\scalebox{0.8}{$\pm$0.17}       &88.40\scalebox{0.8}{$\pm$0.06}    &87.52     \\
prompt-p     &73.98\scalebox{0.8}{$\pm$0.56}    &89.22\scalebox{0.8}{$\pm$0.22}  &\underline{98.62}\scalebox{0.8}{$\pm$0.21} &\underline{90.01}\scalebox{0.8}{$\pm$0.21}   &\underline{87.96}   \\
prompt-a   &74.75\scalebox{0.8}{$\pm$0.86} &\textbf{89.88}\scalebox{0.8}{$\pm$0.28}    &98.57\scalebox{0.8}{$\pm$0.09}    &\textbf{91.79}\scalebox{0.8}{$\pm$0.19}    &\textbf{88.75}     \\
\midrule
param     &72.98\scalebox{0.8}{$\pm$0.05}       &89.44\scalebox{0.8}{$\pm$0.05}      &98.59\scalebox{0.8}{$\pm$0.11}      &88.04\scalebox{0.8}{$\pm$0.12}    &87.26     \\
rma       &74.31\scalebox{0.8}{$\pm$0.45}        &89.52\scalebox{0.8}{$\pm$0.31}      &98.61\scalebox{0.8}{$\pm$0.05}      &88.35\scalebox{0.8}{$\pm$0.24}    &87.70   \\
abd       &74.18\scalebox{0.8}{$\pm$0.20}       &89.59\scalebox{0.8}{$\pm$0.20}      &98.57\scalebox{0.8}{$\pm$0.05}       &88.37\scalebox{0.8}{$\pm$0.12}    &87.68   \\
abd+rma   &74.75\scalebox{0.8}{$\pm$0.27}       &89.52\scalebox{0.8}{$\pm$0.77}      &98.62\scalebox{0.8}{$\pm$0.21}       &88.51\scalebox{0.8}{$\pm$0.18}    &87.85   \\
abd+syn   &73.04\scalebox{0.8}{$\pm$0.28}       &88.90\scalebox{0.8}{$\pm$1.10}      &98.40\scalebox{0.8}{$\pm$0.04}       &88.48\scalebox{0.8}{$\pm$0.16}    &87.48   \\
r+a+s  &\textbf{74.91}\scalebox{0.8}{$\pm$0.39}  &\underline{89.72}\scalebox{0.8}{$\pm$0.22}     &\textbf{98.67}\scalebox{0.8}{$\pm$0.18}   &88.52\scalebox{0.8}{$\pm$0.14}  &\underline{87.96}\\
\bottomrule
\end{tabular}
}
\end{sc}
\end{small}
\end{center}
\vskip -0.1in
\label{ViT-part}
\end{table*}

In the scenario where only the head and the CLS token are tuned \ref{ViT-part}, we can observe that although adding HMA improves upon the backbone, its performance falls short of SOTA prompt tuning \cite{sandler2022fine}. We attribute this to the fact that HMA only adjusts the final feature space, and when most model parameters are fixed, its impact is weaker compared to the method of independently incorporating soft prompts at each layer. However, as demonstrated in the main text, HMA can be effectively combined with prompt tuning to achieve superior performance compared to prompt tuning alone.

\subsection{Algorithm}

\subsubsection{Real Memory Augmentation (RMA)}
\vskip -0.1in

$\textbf{Reading.}$ We utilize Multi-head attention to extract information from the memory buffer, implicitly providing cross-batch information for subsequent SMA. To achieve this, we first transform $M_{\text{buffer}}$ from $\mathbb{R}^{m_1\times d}$ to $\mathbb{R}^{1 \times m_1\times d}$ and $C^1$ from $\mathbb{R}^{bs \times d}$ to $\mathbb{R}^{bs\times 1 \times d}$. Then, we repeat $M_{\text{buffer}}$ $bs$ times in the first dimension to obtain $M_{\text{buffer}}'\in \mathbb{R}^{bs \times m_1\times d}$. Subsequently, we concatenate $M_{\text{buffer}}'$ and $C^1$ along the second dimension, resulting in $C^1_{\text{aug}}\in \mathbb{R}^{bs\times 1+m_1\times d)}$. RMA computes the output feature $C^2$ by:
\begin{equation}
C^2=\text{MHA}(C^1, C_{\text{aug}}^{1},C^1_{\text{aug}})\in \mathbb{R}^{bs\times 1 \times d}
\end{equation}
\vskip -0.05in

$\textbf{Writing.}$  We insert new aggregated input features $[E_{m}(X), E_{\text{label}}(Y)]\in \mathbb{R}^{bs\times d}$ into the memory queue, and update the momentum encoder momentum update rule as in MoCo: $E_{m}^{t+1}=\lambda E_{m}^t + (1 - \lambda) E^{t+1}$, in which $E^{t+1}$ is the SGD-updated encoder from $E^{t}$. 

\subsubsection{Synthetic Memory Augmentation (SMA)}
\label{3.3}
\vskip -0.1in

For each category $i \in [n]$, we initialize $m_2$ learnable memory slots.
Each memory slot is in $\mathbb{R}^{d_1}$ and has corresponding label $y_i$.
With the label embedder $E_{\text{emb}}$, we map the label $y_i$ to the label embedding $e_{\text{syn}, i} \in \mathbb{R}^{d_2}$.
By concatenating all learnable memory slots and label embeddings, we obtain a learnable memory $M_{\text{SMA}} \in \mathbb{R}^{(n * m_2) \times (d_1 + d_2)}$. 
We reshape $C^2$ and $M_{\text{SMA}}$, and concatenate them to obtain $C^2_{\text{aug}} \in \mathbb{R}^{1\times (bs+m_2*n)\times d}$. The SMA output features are obtained by
\begin{equation}
C^3=\text{MHA}(C^2, C^2_{\text{aug}}, C^2_{\text{aug}})\in \mathbb{R}^{1\times bs\times d}.
\end{equation}

Finally, we use a linear projection head to compute the logits from $C^3$. During the training phase, the learnable memory slots are updated through gradient updates, while the label embeddings are regenerated using the updated embedder $E_{\text{emb}}$.

\begin{algorithm}[ht]
    \caption{Heterogeneous Memory Augmentation}
    \label{alg:HMA}
    \begin{algorithmic}[1]
    \REQUIRE data $\mathcal{D}$, encoder $E$, momentum encoder $E_m$, label embedder $E_{\text{label}}$, class number $n$
    \REQUIRE batch size $bs$; buffer size $m_1$, synthetic memory size $m_2$; feature and label embedding dimension $d_1, d_2$
    \STATE initialize empty real memory buffer $M_{\text{buffer}}=\{\}$, synthetic memory $M_{\text{syn}}$
    \FOR{$(X, Y) \in D$}
        \STATE Get input feature $h^1=E(X)$ and unknown embedding $e=E_{\text{label}}(\textbf{n+1})$
        \STATE Read from $M_{\text{buffer}}$ with cross attention according to (\ref{eqn:RMA})
        \STATE Read from $M_{\text{syn}}$ with attention between datapoints according to (\ref{eqn:SMA})
        \STATE Predict with augmented embeddings
        \IF{is training}
            \STATE Update $M_{\text{buffer}}$ with $[E_{m}(X), E_{label}(Y)]$
            \STATE Compute loss and update $E,E_{\text{label}}, M_{\text{syn}}$ and attention weights with gradients
            \STATE Update $E_m=\gamma E_m+(1-\gamma)E$
        \ENDIF
    \ENDFOR
    \end{algorithmic}
\end{algorithm}

\subsection{Societal Impacts}

Due to its lack of additional training targets and stages, as well as its compatibility with various backbone architectures without modifications, HMA can be easily combined with numerous backbones. We believe that HMA has the potential to play a significant role in practical applications of deep learning.

\end{document}